# Graph neural networks: A review of methods and applications

Jie Zhou [a,1], Ganqu Cui [a,1], Shengding Hu [a], Zhengyan Zhang [a], Cheng Yang [b], Zhiyuan Liu [a,*], Lifeng Wang [c], Changcheng Li [c], Maosong Sun [a]

[a] *Department of Computer Science and Technology, Tsinghua University, Beijing, China*
[b] *School of Computer Science, Beijing University of Posts and Telecommunications, China*
[c] *Tencent Incorporation, Shenzhen, China*

ARTICLE INFO

ABSTRACT

Lots of learning tasks require dealing with graph data which contains rich relation information among elements. Modeling physics systems, learning molecular fingerprints, predicting protein interface, and classifying diseases demand a model to learn from graph inputs. In other domains such as learning from non-structural data like texts and images, reasoning on extracted structures (like the dependency trees of sentences and the scene graphs of images) is an important research topic which also needs graph reasoning models. Graph neural networks (GNNs) are neural models that capture the dependence of graphs via message passing between the nodes of graphs. In recent years, variants of GNNs such as graph convolutional network (GCN), graph attention network (GAT), graph recurrent network (GRN) have demonstrated ground-breaking performances on many deep learning tasks. In this survey, we propose a general design pipeline for GNN models and discuss the variants of each component, systematically categorize the applications, and propose four open problems for future research.

## 1. Introduction

Graphs are a kind of data structure which models a set of objects (nodes) and their relationships (edges). Recently, researches on analyzing graphs with machine learning have been receiving more and more attention because of the great expressive power of graphs, i.e. graphs can be used as denotation of a large number of systems across various areas including social science (social networks (Wu et al., 2020), natural science (physical systems (Sanchez et al., 2018; Battaglia et al., 2016) and protein-protein interaction networks (Fout et al., 2017)), knowledge graphs (Hamaguchi et al., 2017) and many other research areas (Khalil et al., 2017). As a unique non-Euclidean data structure for machine learning, graph analysis focuses on tasks such as node classification, link prediction, and clustering. Graph neural networks (GNNs) are deep learning based methods that operate on graph domain. Due to its convincing performance, GNN has become a widely applied graph analysis method recently. In the following paragraphs, we will illustrate the fundamental motivations of graph neural networks.

The first motivation of GNNs roots in the long-standing history of neural networks for graphs. In the nineties, Recursive Neural Networks are first utilized on directed acyclic graphs (Sperduti and Starita, 1997; Frasconi et al., 1998). Afterwards, Recurrent Neural Networks and Feedforward Neural Networks are introduced into this literature respectively in (Scarselli et al., 2009) and (Micheli, 2009) to tackle cycles. Although being successful, the universal idea behind these methods is building state transition systems on graphs and iterate until convergence, which constrained the extendability and representation ability. Recent advancement of deep neural networks, especially convolutional neural networks (CNNs) (LeCun et al., 1998) result in the rediscovery of GNNs. CNNs have the ability to extract multi-scale localized spatial features and compose them to construct highly expressive representations, which led to breakthroughs in almost all machine learning areas and started the new era of deep learning (LeCun et al., 2015). The keys of CNNs are local connection, shared weights and the use of multiple layers (LeCun et al., 2015). These are also of great importance in solving problems on graphs. However, CNNs can only operate on regular Euclidean data like images (2D grids) and texts (1D sequences) while these data structures can be regarded as instances of graphs. Therefore, it is straightforward to generalize CNNs on graphs. As shown in Fig. 1, it is hard to define localized convolutional filters and pooling operators, which hinders the transformation of CNN from Euclidean domain to non-Euclidean domain. Extending deep neural models to non-Euclidean domains, which is generally referred to as geometric deep learning, has been an emerging research area (Bronstein et al., 2017). Under this umbrella term, deep learning on graphs receives enormous attention.

* Corresponding author.
*E-mail addresses:* zhoujie18@mails.tsinghua.edu.cn (J. Zhou), cgq19@mails.tsinghua.edu.cn (G. Cui), hsd20@mails.tsinghua.edu.cn (S. Hu), zy-z19@mails.tsinghua.edu.cn (Z. Zhang), albertyang33@gmail.com (C. Yang), liuzy@tsinghua.edu.cn (Z. Liu), fandywang@tencent.com (L. Wang), harrychli@tencent.com (C. Li), sms@tsinghua.edu.cn (M. Sun).
[1] indicates equal contribution.

The other motivation comes from *graph representation learning* (Cui et al., 2018a; Hamilton et al., 2017b; Zhang et al., 2018a; Cai et al., 2018; Goyal and Ferrara, 2018), which learns to represent graph nodes, edges or subgraphs by low-dimensional vectors. In the field of graph analysis, traditional machine learning approaches usually rely on hand engineered features and are limited by its inflexibility and high cost. Following the idea of *representation learning* and the success of word embedding (Mikolov et al., 2013), DeepWalk (Perozzi et al., 2014), regarded as the first graph embedding method based on representation learning, applies SkipGram model (Mikolov et al., 2013) on the generated random walks. Similar approaches such as node2vec (Grover and Leskovec, 2016), LINE (Tang et al., 2015) and TADW (Yang et al., 2015) also achieved breakthroughs. However, these methods suffer from two severe drawbacks (Hamilton et al., 2017b). First, no parameters are shared between nodes in the encoder, which leads to computationally inefficiency, since it means the number of parameters grows linearly with the number of nodes. Second, the direct embedding methods lack the ability of generalization, which means they cannot deal with dynamic graphs or generalize to new graphs.

Based on CNNs and graph embedding, variants of graph neural networks (GNNs) are proposed to collectively aggregate information from graph structure. Thus they can model input and/or output consisting of elements and their dependency.

There exists several comprehensive reviews on graph neural networks. Bronstein et al. (2017) provide a thorough review of geometric deep learning, which presents its problems, difficulties, solutions, applications and future directions. Zhang et al. (2019a) propose another comprehensive overview of graph convolutional networks. However, they mainly focus on convolution operators defined on graphs while we investigate other computation modules in GNNs such as skip connections and pooling operators.

Papers by Zhang et al. (2018b), Wu et al. (2019a), Chami et al. (2020) are the most up-to-date survey papers on GNNs and they mainly focus on models of GNN. Wu et al. (2019a) categorize GNNs into four groups: recurrent graph neural networks, convolutional graph neural networks, graph autoencoders, and spatial-temporal graph neural networks. Zhang et al. (2018b) give a systematic overview of different graph deep learning methods and Chami et al. (2020) propose a Graph Encoder Decoder Model to unify network embedding and graph neural network models. Our paper provides a different taxonomy with them and we mainly focus on classic GNN models. Besides, we summarize variants of GNNs for different graph types and also provide a detailed summary of GNNs' applications in different domains.

There have also been several surveys focusing on some specific graph learning fields. Sun et al. (2018) and Chen et al. (2020a) give detailed overviews for adversarial learning methods on graphs, including graph data attack and defense. Lee et al. (2018a) provide a review over graph attention models. The paper proposed by Yang et al. (2020) focuses on heterogeneous graph representation learning, where nodes or edges are of multiple types. Huang et al. (2020) review over existing GNN models for dynamic graphs. Peng et al. (2020) summarize graph embeddings methods for combinatorial optimization. We conclude GNNs for heterogeneous graphs, dynamic graphs and combinatorial optimization in Section 4.2, Section 4.3, and Section 8.1.6 respectively.

In this paper, we provide a thorough review of different graph neural network models as well as a systematic taxonomy of the applications. To summarize, our contributions are:

- We provide a detailed review over existing graph neural network models. We present a general design pipeline and discuss the variants of each module. We also introduce researches on theoretical and empirical analyses of GNN models.
- We systematically categorize the applications and divide the applications into structural scenarios and non-structural scenarios. We present several major applications and their corresponding methods for each scenario.
- We propose four open problems for future research. We provide a thorough analysis of each problem and propose future research directions.

The rest of this survey is organized as follows. In Section 2, we present a general GNN design pipeline. Following the pipeline, we discuss each step in detail to review GNN model variants. The details are included in Section 3 to Section 6. In Section 7, we revisit research works over theoretical and empirical analyses of GNNs. In Section 8, we introduce several major applications of graph neural networks applied to structural scenarios, non-structural scenarios and other scenarios. In Section 9, we propose four open problems of graph neural networks as well as several future research directions. And finally, we conclude the survey in Section 10.

## 2. General design pipeline of GNNs

In this paper, we introduce models of GNNs in a designer view. We first present the general design pipeline for designing a GNN model in this section. Then we give details of each step such as selecting computational modules, considering graph type and scale, and designing loss function in Section 3, 4, and 5, respectively. And finally, we use an example to illustrate the design process of GNN for a specific task in Section 6.

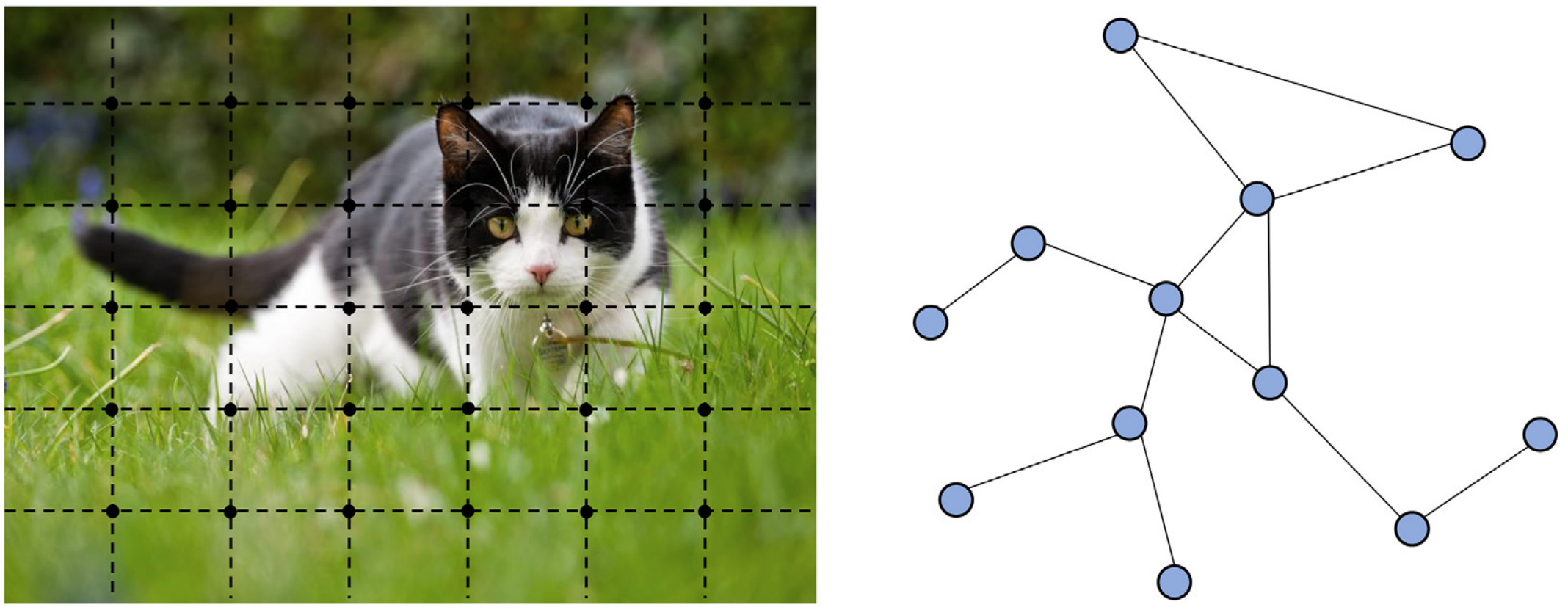

**Fig. 1.** Left: image in Euclidean space. Right: graph in non-Euclidean space.

In later sections, we denote a graph as $G = (V, E)$, where $|V| = N$ is the number of nodes in the graph and $|E| = N^e$ is the number of edges. $\mathbf{A} \in \mathbb{R}^{N \times N}$ is the adjacency matrix. For graph representation learning, we use $\mathbf{h}_v$ and $\mathbf{o}_v$ as the hidden state and output vector of node $v$. The detailed descriptions of the notations could be found in Table 1.

In this section, we present the general design pipeline of a GNN model for a specific task on a specific graph type. Generally, the pipeline contains four steps: (1) find graph structure, (2) specify graph type and scale, (3) design loss function and (4) build model using computational modules. We give general design principles and some background knowledge in this section. The design details of these steps are discussed in later sections.

### 2.1. Find graph structure

At first, we have to find out the graph structure in the application. There are usually two scenarios: structural scenarios and non-structural scenarios. In structural scenarios, the graph structure is explicit in the applications, such as applications on molecules, physical systems, knowledge graphs and so on. In non-structural scenarios, graphs are implicit so that we have to first build the graph from the task, such as building a fully-connected "word" graph for text or building a scene graph for an image. After we get the graph, the later design process attempts to find an optimal GNN model on this specific graph.

### 2.2. Specify graph type and scale

After we get the graph in the application, we then have to find out the graph type and its scale.

Graphs with complex types could provide more information on nodes and their connections. Graphs are usually categorized as:

- **Directed/Undirected Graphs.** Edges in directed graphs are all directed from one node to another, which provide more information than undirected graphs. Each edge in undirected graphs can also be regarded as two directed edges.
- **Homogeneous/Heterogeneous Graphs.** Nodes and edges in homogeneous graphs have same types, while nodes and edges have different types in heterogeneous graphs. Types for nodes and edges play important roles in heterogeneous graphs and should be further considered.
- **Static/Dynamic Graphs.** When input features or the topology of the graph vary with time, the graph is regarded as a dynamic graph. The time information should be carefully considered in dynamic graphs.

Note these categories are orthogonal, which means these types can be combined, e.g. one can deal with a dynamic directed heterogeneous graph. There are also several other graph types designed for different tasks such as hypergraphs and signed graphs. We will not enumerate all types here but the most important idea is to consider the additional information provided by these graphs. Once we specify the graph type, the additional information provided by these graph types should be further considered in the design process.

As for the graph scale, there is no clear classification criterion for "small" and "large" graphs. The criterion is still changing with the development of computation devices (e.g. the speed and memory of GPUs). In this paper, when the adjacency matrix or the graph Laplacian of a graph (the space complexity is $O(n^2)$) cannot be stored and processed by the device, then we regard the graph as a large-scale graph and then some sampling methods should be considered.

**Table 1**
Notations used in this paper.

| Notations | Descriptions |
|---|---|
| $\mathbb{R}^m$ | $m$-dimensional Euclidean space |
| $a, \mathbf{a}, \mathbf{A}$ | Scalar, vector and matrix |
| $\mathbf{A}^T$ | Matrix transpose |
| $\mathbf{I}_N$ | Identity matrix of dimension $N$ |
| $\mathbf{g}_w \star \mathbf{x}$ | Convolution of $\mathbf{g}_w$ and $\mathbf{x}$ |
| $N$, $N^v$ | Number of nodes in the graph |
| $N^e$ | Number of edges in the graph |
| $\mathcal{N}_v$ | Neighborhood set of node $v$ |
| $\mathbf{a}_v^t$ | Vector $\mathbf{a}$ of node $v$ at time step $t$ |
| $\mathbf{h}_v$ | Hidden state of node $v$ |
| $\mathbf{h}_v^t$ | Hidden state of node $v$ at time step $t$ |
| $\mathbf{o}_v^t$ | Output of node $v$ at time step $t$ |
| $\mathbf{e}_{vw}$ | Features of edge from node $v$ to $w$ |
| $\mathbf{e}_k$ | Features of edge with label $k$ |
| $\mathbf{W}^i, \mathbf{U}^i, \mathbf{W}^o, \mathbf{U}^o$ | Matrices for computing $\mathbf{i}, \mathbf{o}$. |
| $\mathbf{b}^i, \mathbf{b}^o$ | Vectors for computing $\mathbf{i}, \mathbf{o}$ |
| $\rho$ | An alternative non-linear function |
| $\sigma$ | The logistic sigmoid function |
| tanh | The hyperbolic tangent function |
| LeakyReLU | The LeakyReLU function |
| $\odot$ | Element-wise multiplication operation |
| $\Vert$ | Vector concatenation |

### 2.3. Design loss function

In this step we should design the loss function based on our task type and the training setting.

For graph learning tasks, there are usually three kinds of tasks:

- **Node-level** tasks focus on nodes, which include node classification, node regression, node clustering, etc. Node classification tries to categorize nodes into several classes, and node regression predicts a continuous value for each node. Node clustering aims to partition the nodes into several disjoint groups, where similar nodes should be in the same group.
- **Edge-level** tasks are edge classification and link prediction, which require the model to classify edge types or predict whether there is an edge existing between two given nodes.
- **Graph-level** tasks include graph classification, graph regression, and graph matching, all of which need the model to learn graph representations.

From the perspective of supervision, we can also categorize graph learning tasks into three different training settings:

- **Supervised setting** provides labeled data for training.
- **Semi-supervised setting** gives a small amount of labeled nodes and a large amount of unlabeled nodes for training. In the test phase, the **transductive setting** requires the model to predict the labels of the given unlabeled nodes, while the **inductive setting** provides new unlabeled nodes from the same distribution to infer. Most node and edge classification tasks are semi-supervised. Most recently, a mixed transductive-inductive scheme is undertaken by Wang and Leskovec (2020) and Rossi et al. (2018), craving a new path towards the mixed setting.
- **Unsupervised setting** only offers unlabeled data for the model to find patterns. Node clustering is a typical unsupervised learning task.

With the task type and the training setting, we can design a specific loss function for the task. For example, for a node-level semi-supervised classification task, the cross-entropy loss can be used for the labeled nodes in the training set.

### 2.4. Build model using computational modules

Finally, we can start building the model using the computational modules. Some commonly used computational modules are:

- **Propagation Module.** The propagation module is used to propagate information between nodes so that the aggregated information could capture both feature and topological information. In propagation modules, the **convolution operator** and **recurrent operator** are usually used to aggregate information from neighbors while the **skip connection** operation is used to gather information from historical representations of nodes and mitigate the over-smoothing problem.
- **Sampling Module.** When graphs are large, sampling modules are usually needed to conduct propagation on graphs. The sampling module is usually combined with the propagation module.
- **Pooling Module.** When we need the representations of high-level subgraphs or graphs, pooling modules are needed to extract information from nodes.

With these computation modules, a typical GNN model is usually built by combining them. A typical architecture of the GNN model is illustrated in the middle part of Fig. 2 where the convolutional operator, recurrent operator, sampling module and skip connection are used to propagate information in each layer and then the pooling module is added to extract high-level information. These layers are usually stacked to obtain better representations. Note this architecture can generalize most GNN models while there are also exceptions, for example, NDCN (Zang and Wang, 2020) combines ordinary differential equation systems (ODEs) and GNNs. It can be regarded as a continuous-time GNN model which integrates GNN layers over continuous time without propagating through a discrete number of layers.

An illustration of the general design pipeline is shown in Fig. 2. In later sections, we first give the existing instantiations of computational modules in Section 3, then introduce existing variants which consider different graph types and scale in Section 4. Then we survey on variants designed for different training settings in Section 5. These sections correspond to details of step (4), step (2), and step (3) in the pipeline. And finally, we give a concrete design example in Section 6.

## 3. Instantiations of computational modules

In this section we introduce existing instantiations of three computational modules: propagation modules, sampling modules and pooling modules. We introduce three sub-components of propagation modules: convolution operator, recurrent operator and skip connection in Section 3.1, 3.2, and 3.3 respectively. Then we introduce sampling modules and pooling modules in Section 3.4 and 3.5. An overview of computational modules is shown in Fig. 3.

### 3.1. Propagation modules - convolution operator

Convolution operators that we introduce in this section are the mostly used propagation operators for GNN models. The main idea of convolution operators is to generalize convolutions from other domain to the graph domain. Advances in this direction are often categorized as spectral approaches and spatial approaches.

#### 3.1.1. Spectral approaches

Spectral approaches work with a spectral representation of the graphs. These methods are theoretically based on graph signal processing (Shuman et al., 2013) and define the convolution operator in the spectral domain.

In spectral methods, a graph signal $\mathbf{x}$ is firstly transformed to the spectral domain by the graph Fourier transform $\mathscr{F}$, then the convolution operation is conducted. After the convolution, the resulted signal is transformed back using the inverse graph Fourier transform $\mathscr{F}^{-1}$. These transforms are defined as:

$$\begin{aligned}\mathscr{F}(\mathbf{x}) &= \mathbf{U}^T\mathbf{x},\\ \mathscr{F}^{-1}(\mathbf{x}) &= \mathbf{U}\mathbf{x}.\end{aligned} \tag{1}$$

Here $\mathbf{U}$ is the matrix of eigenvectors of the normalized graph Laplacian $\mathbf{L} = \mathbf{I}_N - \mathbf{D}^{-\frac{1}{2}}\mathbf{A}\mathbf{D}^{-\frac{1}{2}}$ ($\mathbf{D}$ is the degree matrix and $\mathbf{A}$ is the adjacency matrix of the graph). The normalized graph Laplacian is real symmetric positive semidefinite, so it can be factorized as $\mathbf{L} = \mathbf{U}\mathbf{\Lambda}\mathbf{U}^T$ (where $\mathbf{\Lambda}$ is a diagonal matrix of the eigenvalues). Based on the convolution theorem (Mallat, 1999), the convolution operation is defined as:

$$\begin{aligned}\mathbf{g} \star \mathbf{x} &= \mathscr{F}^{-1}(\mathscr{F}(\mathbf{g}) \odot \mathscr{F}(\mathbf{x}))\\ &= \mathbf{U}(\mathbf{U}^T\mathbf{g} \odot \mathbf{U}^T\mathbf{x}),\end{aligned} \tag{2}$$

where $\mathbf{U}^T\mathbf{g}$ is the filter in the spectral domain. If we simplify the filter by

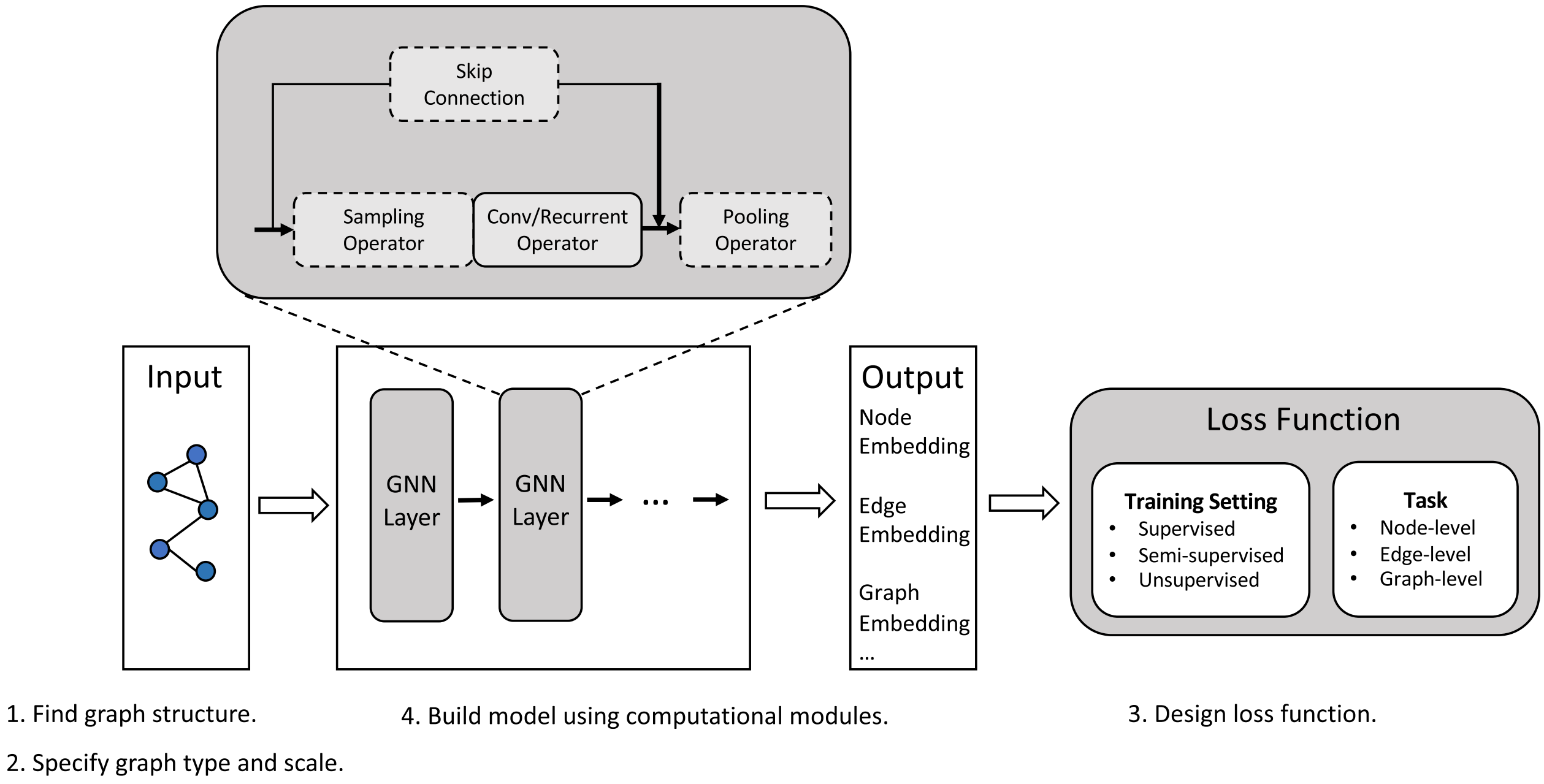

**Fig. 2.** The general design pipeline for a GNN model.

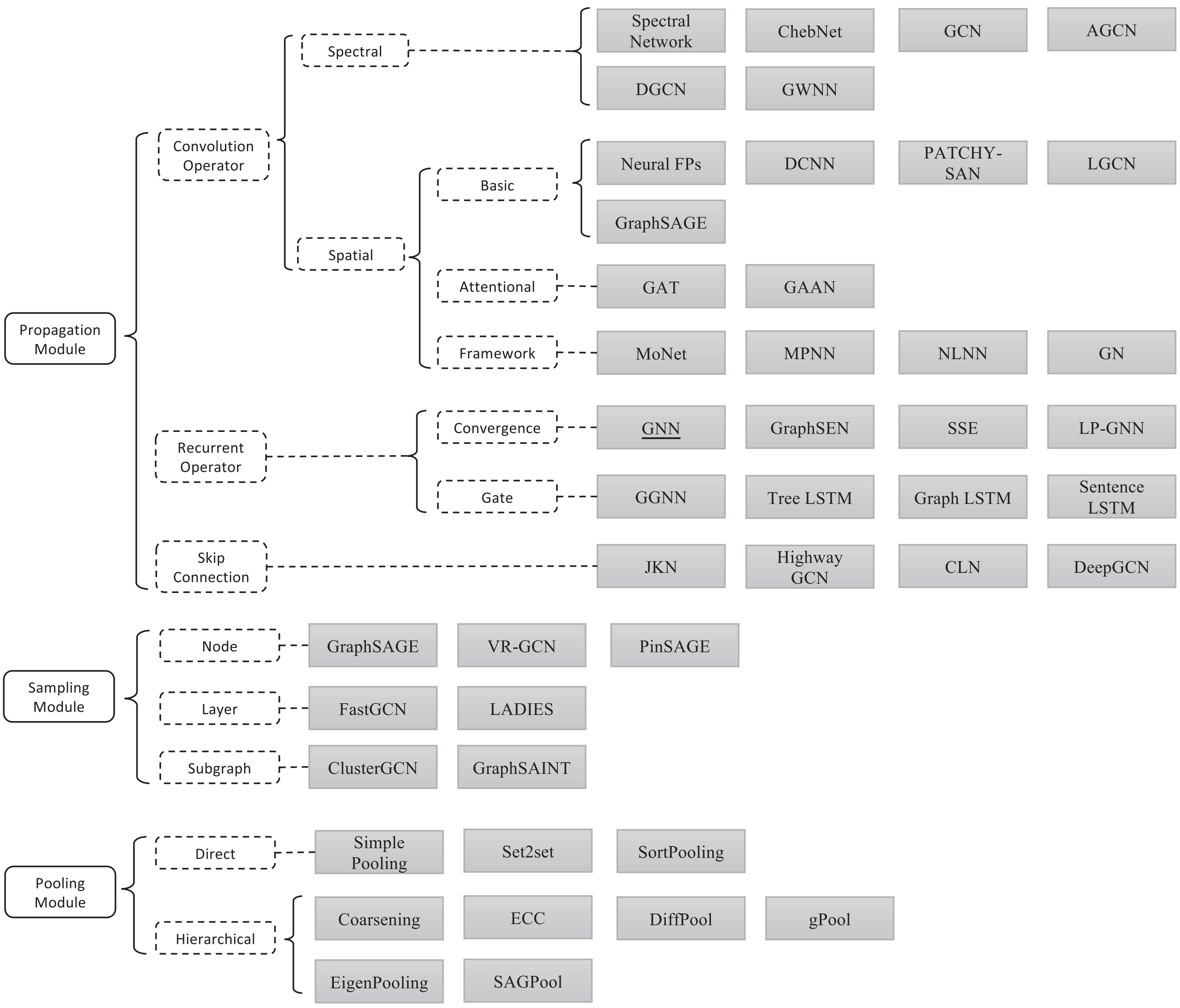

**Fig. 3.** An overview of computational modules.

using a learnable diagonal matrix $\mathbf{g}_w$, then we have the basic function of the spectral methods:

$$\mathbf{g}_w \star \mathbf{x} = \mathbf{U}\mathbf{g}_w\mathbf{U}^T\mathbf{x}. \tag{3}$$

Next we introduce several typical spectral methods which design different filters $\mathbf{g}_w$.

*Spectral Network.* Spectral network (Bruna et al., 2014) uses a learnable diagonal matrix as the filter, that is $\mathbf{g}_w = \text{diag}(\mathbf{w})$, where $\mathbf{w} \in \mathbb{R}^N$ is the parameter. However, this operation is computationally inefficient and the filter is non-spatially localized. Henaff et al. (2015) attempt to make the spectral filters spatially localized by introducing a parameterization with smooth coefficients.

*ChebNet.* Hammond et al. (2011) suggest that $\mathbf{g}_w$ can be approximated by a truncated expansion in terms of Chebyshev polynomials $\mathbf{T}_k(x)$ up to $K^{th}$ order. Defferrard et al. (2016) propose the ChebNet based on this theory. Thus the operation can be written as:

$$\mathbf{g}_w \star \mathbf{x} \approx \sum_{k=0}^{K} w_k \mathbf{T}_k\left(\tilde{\mathbf{L}}\right)\mathbf{x}, \tag{4}$$

where $\tilde{\mathbf{L}} = \frac{2}{\lambda_{max}}\mathbf{L} - \mathbf{I}_N$, $\lambda_{max}$ denotes the largest eigenvalue of $\mathbf{L}$. The range of the eigenvalues in $\tilde{\mathbf{L}}$ is [-1, 1]. $\mathbf{w} \in \mathbb{R}^K$ is now a vector of Chebyshev coefficients. The Chebyshev polynomials are defined as $\mathbf{T}_k(\mathbf{x}) = 2\mathbf{x}\mathbf{T}_{k-1}(\mathbf{x}) - \mathbf{T}_{k-2}(\mathbf{x})$, with $\mathbf{T}_0(\mathbf{x}) = 1$ and $\mathbf{T}_1(\mathbf{x}) = \mathbf{x}$. It can be observed that the operation is $K$-localized since it is a $K^{th}$-order polynomial in the Laplacian. Defferrard et al. (2016) use this $K$-localized convolution to define a convolutional neural network which could remove the need to compute the eigenvectors of the Laplacian.

*GCN.* Kipf and Welling (2017) simplify the convolution operation in Eq. (4) with $K = 1$ to alleviate the problem of overfitting. They further assume $\lambda_{max} \approx 2$ and simplify the equation to

$$\mathbf{g}_w \star \mathbf{x} \approx w_0\mathbf{x} + w_1(\mathbf{L} - \mathbf{I}_N)\mathbf{x} = w_0\mathbf{x} - w_1\mathbf{D}^{-\frac{1}{2}}\mathbf{A}\mathbf{D}^{-\frac{1}{2}}\mathbf{x} \tag{5}$$

with two free parameters $w_0$ and $w_1$. With parameter constraint $w = w_0 = -w_1$, we can obtain the following expression:

$$\mathbf{g}_w \star \mathbf{x} \approx w\left(\mathbf{I}_N + \mathbf{D}^{-\frac{1}{2}}\mathbf{A}\mathbf{D}^{-\frac{1}{2}}\right)\mathbf{x}. \tag{6}$$

GCN further introduces a *renormalization trick* to solve the exploding/vanishing gradient problem in Eq. (6): $\mathbf{I}_N + \mathbf{D}^{-\frac{1}{2}}\mathbf{A}\mathbf{D}^{-\frac{1}{2}} \to \tilde{\mathbf{D}}^{-\frac{1}{2}}\tilde{\mathbf{A}}\tilde{\mathbf{D}}^{-\frac{1}{2}}$, with $\tilde{\mathbf{A}} = \mathbf{A} + \mathbf{I}_N$ and $\tilde{\mathbf{D}}_{ii} = \sum_j \tilde{\mathbf{A}}_{ij}$. Finally, the compact form of GCN is defined as:

$$\mathbf{H} = \tilde{\mathbf{D}}^{-\frac{1}{2}}\tilde{\mathbf{A}}\tilde{\mathbf{D}}^{-\frac{1}{2}}\mathbf{X}\mathbf{W}, \tag{7}$$

where $\mathbf{X} \in \mathbb{R}^{N\times F}$ is the input matrix, $\mathbf{W} \in \mathbb{R}^{F\times F'}$ is the parameter and $\mathbf{H} \in \mathbb{R}^{N\times F'}$ is the convolved matrix. $F$ and $F'$ are the dimensions of the input and the output, respectively. Note that GCN can also be regarded as a spatial method that we will discuss later.

*AGCN.* All of these models use the original graph structure to denote relations between nodes. However, there may have implicit relations between different nodes. The Adaptive Graph Convolution Network (AGCN) is proposed to learn the underlying relations (Li et al., 2018a). AGCN learns a "residual" graph Laplacian and add it to the original Laplacian matrix. As a result, it is proven to be effective in several graph-structured datasets.

*DGCN.* The dual graph convolutional network (DGCN) (Zhuang and Ma, 2018) is proposed to jointly consider the local consistency and global consistency on graphs. It uses two convolutional networks to capture the local and global consistency and adopts an unsupervised loss to ensemble them. The first convolutional network is the same as Eq. (7), and the second network replaces the adjacency matrix with positive pointwise mutual information (PPMI) matrix:

$$\mathbf{H}' = \rho\left(\mathbf{D}_P^{-\frac{1}{2}}\mathbf{A}_P\mathbf{D}_P^{-\frac{1}{2}}\mathbf{H}\mathbf{W}\right), \tag{8}$$

where $\mathbf{A}_P$ is the PPMI matrix and $\mathbf{D}_P$ is the diagonal degree matrix of $\mathbf{A}_P$.

*GWNN.* Graph wavelet neural network (GWNN) (Xu et al., 2019a) uses the graph wavelet transform to replace the graph Fourier transform. It has several advantages: (1) graph wavelets can be fastly obtained without matrix decomposition; (2) graph wavelets are sparse and localized thus the results are better and more explainable. GWNN outperforms several spectral methods on the semi-supervised node classification task.

*AGCN* and *DGCN* try to improve spectral methods from the perspective of augmenting graph Laplacian while *GWNN* replaces the Fourier transform. In conclusion, spectral approaches are well theoretically based and there are also several theoretical analyses proposed recently (see Section 7.1.1). However, in almost all of the spectral approaches mentioned above, the learned filters depend on graph structure. That is to say, the filters cannot be applied to a graph with a different structure and those models can only be applied under the "transductive" setting of graph tasks.

### 3.1.2. Basic spatial approaches

Spatial approaches define convolutions directly on the graph based on the graph topology. The major challenge of spatial approaches is defining the convolution operation with differently sized neighborhoods and maintaining the local invariance of CNNs.

*Neural FPs.* Neural FPs (Duvenaud et al., 2015) uses different weight matrices for nodes with different degrees:

$$\begin{aligned}\mathbf{t} &= \mathbf{h}_v^t + \sum_{u\in\mathcal{N}_v}\mathbf{h}_u^t,\\ \mathbf{h}_v^{t+1} &= \sigma\left(\mathbf{t}\mathbf{W}_{|\mathcal{N}_v|}^{t+1}\right),\end{aligned} \tag{9}$$

where $\mathbf{W}_{|\mathcal{N}_v|}^{t+1}$ is the weight matrix for nodes with degree $|\mathcal{N}_v|$ at layer $t+1$. The main drawback of the method is that it cannot be applied to large-scale graphs with more node degrees.

*DCNN.* The diffusion convolutional neural network (DCNN) (Atwood and Towsley, 2016) uses transition matrices to define the neighborhood for nodes. For node classification, the diffusion representations of each node in the graph can be expressed as:

$$\mathbf{H} = f(\mathbf{W}_c \odot \mathbf{P}^*\mathbf{X}) \in \mathbb{R}^{N\times K\times F}, \tag{10}$$

where $\mathbf{X} \in \mathbb{R}^{N\times F}$ is the matrix of input features ($F$ is the dimension). $\mathbf{P}^*$ is an $N \times K \times N$ tensor which contains the power series $\{\mathbf{P}, \mathbf{P}^2, \ldots, \mathbf{P}^K\}$ of matrix $\mathbf{P}$. And $\mathbf{P}$ is the degree-normalized transition matrix from the graphs adjacency matrix $\mathbf{A}$. Each entity is transformed to a diffusion convolutional representation which is a $K \times F$ matrix defined by $K$ hops of graph diffusion over $F$ features. And then it will be defined by a $K \times F$ weight matrix and a non-linear activation function $f$.

*PATCHY-SAN.* The PATCHY-SAN model (Niepert et al., 2016) extracts and normalizes a neighborhood of exactly $k$ nodes for each node. The normalized neighborhood serves as the receptive field in the traditional convolutional operation.

*LGCN.* The learnable graph convolutional network (LGCN) (Gao et al., 2018a) also exploits CNNs as aggregators. It performs max pooling on neighborhood matrices of nodes to get top-k feature elements and then applies 1-D CNN to compute hidden representations.

*GraphSAGE.* GraphSAGE (Hamilton et al., 2017a) is a general inductive framework which generates embeddings by sampling and aggregating features from a node's local neighborhood:

$$\begin{aligned}\mathbf{h}_{\mathcal{N}_v}^{t+1} &= \mathrm{AGG}_{t+1}\left(\{\mathbf{h}_u^t, \forall u \in \mathcal{N}_v\}\right),\\ \mathbf{h}_v^{t+1} &= \sigma\left(\mathbf{W}^{t+1}\cdot\left[\mathbf{h}_v^t \,\|\, \mathbf{h}_{\mathcal{N}_v}^{t+1}\right]\right).\end{aligned} \tag{11}$$

Instead of using the full neighbor set, GraphSAGE uniformly samples a fixed-size set of neighbors to aggregate information. $\mathrm{AGG}_{t+1}$ is the aggregation function and GraphSAGE suggests three aggregators: mean aggregator, LSTM aggregator, and pooling aggregator. GraphSAGE with a mean aggregator can be regarded as an inductive version of GCN while the LSTM aggregator is not permutation invariant, which requires a specified order of the nodes.

### 3.1.3. Attention-based spatial approaches

The attention mechanism has been successfully used in many sequence-based tasks such as machine translation (Bahdanau et al., 2015; Gehring et al., 2017; Vaswani et al., 2017), machine reading (Cheng et al., 2016) and so on. There are also several models which try to generalize the attention operator on graphs (Velickovic et al., 2018; Zhang et al., 2018c). Compared with the operators we mentioned before, attention-based operators assign different weights for neighbors, so that they could alleviate noises and achieve better results.

*GAT.* The graph attention network (GAT) (Velickovic et al., 2018) incorporates the attention mechanism into the propagation step. It computes the hidden states of each node by attending to its neighbors, following a *self-attention* strategy. The hidden state of node $v$ can be obtained by:

$$\begin{aligned}\mathbf{h}_v^{t+1} &= \rho\left(\sum_{u\in\mathcal{N}_v}\alpha_{vu}\mathbf{W}\mathbf{h}_u^t\right),\\ \alpha_{vu} &= \frac{\exp(\mathrm{LeakyReLU}(\mathbf{a}^T[\mathbf{W}\mathbf{h}_v \,\|\, \mathbf{W}\mathbf{h}_u]))}{\sum_{k\in\mathcal{N}_v}\exp(\mathrm{LeakyReLU}(\mathbf{a}^T[\mathbf{W}\mathbf{h}_v \,\|\, \mathbf{W}\mathbf{h}_k]))},\end{aligned} \tag{12}$$

where $\mathbf{W}$ is the weight matrix associated with the linear transformation which is applied to each node, and $\mathbf{a}$ is the weight vector of a single-layer MLP.

Moreover, GAT utilizes the *multi-head attention* used by Vaswani et al. (2017) to stabilize the learning process. It applies $K$ independent attention head matrices to compute the hidden states and then concatenates their features (or computes the average), resulting in the following two

output representations:

$$\mathbf{h}_v^{t+1} = \|_{k=1}^K \sigma\left(\sum_{u\in\mathcal{N}_v}\alpha_{vu}^k \mathbf{W}_k \mathbf{h}_u^t\right),$$
$$\mathbf{h}_v^{t+1} = \sigma\left(\frac{1}{K}\sum_{k=1}^K\sum_{u\in\mathcal{N}_v}\alpha_{vu}^k \mathbf{W}_k \mathbf{h}_u^t\right). \tag{13}$$

Here $\alpha_{ij}^k$ is the normalized attention coefficient computed by the $k$-th attention head. The attention architecture has several properties: (1) the computation of the node-neighbor pairs is parallelizable thus the operation is efficient; (2) it can be applied to graph nodes with different degrees by specifying arbitrary weights to neighbors; (3) it can be applied to the inductive learning problems easily.

*GaAN.* The gated attention network (GaAN) (Zhang et al., 2018c) also uses the multi-head attention mechanism. However, it uses a self-attention mechanism to gather information from different heads to replace the average operation of GAT.

#### 3.1.4. General frameworks for spatial approaches

Apart from different variants of spatial approaches, several general frameworks are proposed aiming to integrate different models into one single framework. Monti et al. (2017) propose the mixture model network (MoNet), which is a general spatial framework for several methods defined on graphs or manifolds. Gilmer et al. (2017) propose the message passing neural network (MPNN), which uses message passing functions to unify several variants. Wang et al. (2018a) propose the non-local neural network (NLNN) which unifies several "self-attention"-style methods (Hoshen, 2017; Vaswani et al., 2017; Velickovic et al., 2018). Battaglia et al. (2018) propose the graph network (GN). It defines a more general framework for learning node-level, edge-level and graph-level representations.

*MoNet.* Mixture model network (MoNet) (Monti et al., 2017) is a spatial framework that try to unifies models for non-euclidean domains, including CNNs for manifold and GNNs. The Geodesic CNN (GCNN) (Masci et al., 2015) and Anisotropic CNN (ACNN) (Boscaini et al., 2016) on manifolds or GCN (Kipf and Welling, 2017) and DCNN (Atwood and Towsley, 2016) on graphs can be formulated as particular instances of MoNet. In MoNet, each point on a manifold or each vertex on a graph, denoted by $v$, is regarded as the origin of a pseudo-coordinate system. The neighbors $u\in\mathcal{N}_v$ are associated with pseudo-coordinates $\mathbf{u}(v,u)$. Given two functions $f,g$ defined on the vertices of a graph (or points on a manifold), the convolution operator in MoNet is defined as:

$$(f\star g) = \sum_{j=1}^J g_j D_j(v)f,$$
$$D_j(v)f = \sum_{u\in\mathcal{N}_v} w_j(\mathbf{u}(v,u))f(u). \tag{14}$$

Here $w_1(\mathbf{u}),\dots,w_J(\mathbf{u})$ are the functions assigning weights for neighbors according to their pseudo-coordinates. Thus the $D_j(v)f$ is the aggregated values of the neighbors' functions. By defining different $\mathbf{u}$ and $\mathbf{w}$, MoNet can instantiate several methods. For GCN, the function $f,g$ map the nodes to their features, the pseudo-coordinate for $(v,u)$ is $\mathbf{u}(v,u) = (|\mathcal{N}_v|,|\mathcal{N}_u|)$, $J=1$ and $w_1(\mathbf{u}(v,u)) = \frac{1}{\sqrt{|\mathcal{N}_v||\mathcal{N}_u|}}$. In MoNet's own model, the parameters $w_j$ are learnable.

*MPNN.* The message passing neural network (MPNN) (Gilmer et al., 2017) extracts the general characteristics among several classic models. The model contains two phases: a message passing phase and a readout phase. In the message passing phase, the model first uses the message function $M_t$ to aggregate the "message" $\mathbf{m}_v^t$ from neighbors and then uses the update function $U_t$ to update the hidden state $\mathbf{h}_v^t$:

$$\mathbf{m}_v^{t+1} = \sum_{u\in\mathcal{N}_v} M_t\left(\mathbf{h}_v^t, \mathbf{h}_u^t, \mathbf{e}_{vu}\right),$$
$$\mathbf{h}_v^{t+1} = U_t\left(\mathbf{h}_v^t, \mathbf{m}_v^{t+1}\right). \tag{15}$$

Here $\mathbf{e}_{vu}$ represents features of undirected edge $(v,u)$. The readout phase computes a feature vector of the whole graph using the readout function $R$:

$$\hat{\mathbf{y}} = R\left(\left\{\mathbf{h}_v^T \middle| v\in G\right\}\right), \tag{16}$$

where T denotes the total time steps. The message function $M_t$, vertex update function $U_t$ and readout function $R$ may have different settings. Hence the MPNN framework could instantiate several different models via different function settings. Specific settings for different models could be found in (Gilmer et al., 2017).

*NLNN.* The non-local neural network (NLNN) generalizes and extends the classic non-local mean operation (Buades et al., 2005) in computer vision. The non-local operation computes the hidden state at a position as a weighted sum of features at all possible positions. The potential positions can be in space, time or spacetime. Thus the NLNN can be viewed as a unification of different "self-attention"-style methods (Hoshen, 2017; Vaswani et al., 2017; Velickovic et al., 2018).

Following the non-local mean operation (Buades et al., 2005), the generic non-local operation is defined as

$$\mathbf{h}_v^{t+1} = \frac{1}{\mathcal{C}(\mathbf{h}^t)}\sum_{\forall u} f\left(\mathbf{h}_v^t, \mathbf{h}_u^t\right) g\left(\mathbf{h}_u^t\right), \tag{17}$$

where $u$ is the index of all possible positions for position $v$, $f(\mathbf{h}_v^t, \mathbf{h}_u^t)$ computes a scalar between $v$ and $u$ representing the relation between them, $g(\mathbf{h}_u^t)$ denotes a transformation of the input $\mathbf{h}_u^t$ and $\mathcal{C}(\mathbf{h}^t)$ is a normalization factor. Different variants of NLNN can be defined by different $f$ and $g$ settings and more details can be found in the original paper (Buades et al., 2005).

*Graph Network.* The graph network (GN) (Battaglia et al., 2018) is a more general framework compared to others by learning node-level, edge-level and graph level representations. It can unify many variants like MPNN, NLNN, Interaction Networks (Battaglia et al., 2016; Watters et al., 2017), Neural Physics Engine (Chang et al., 2017), CommNet (Sukhbaatar Ferguset al., 2016), structure2vec (Dai et al., 2016; Khalil et al., 2017), GGNN (Li et al., 2016), Relation Network (Raposo et al., 2017; Santoro et al., 2017), Deep Sets (Zaheer et al., 2017), Point Net (Qi et al., 2017a) and so on.

The core computation unit of GN is called the GN block. A GN block defines three update functions and three aggregation functions:

$$\mathbf{e}_k^{t+1} = \varphi^e\left(\mathbf{e}_k^t, \mathbf{h}_{r_k}^t, \mathbf{h}_{s_k}^t, \mathbf{u}^t\right),\ \bar{\mathbf{e}}_v^{t+1} = \rho^{e\to h}\left(\mathbf{E}_v^{t+1}\right),$$
$$\mathbf{h}_v^{t+1} = \varphi^h\left(\bar{\mathbf{e}}_v^{t+1}, \mathbf{h}_v^t, \mathbf{u}^t\right),\ \bar{\mathbf{e}}^{t+1} = \rho^{e\to u}\left(\mathbf{E}^{t+1}\right), \tag{18}$$
$$\mathbf{u}^{t+1} = \varphi^u\left(\bar{\mathbf{e}}^{t+1}, \bar{\mathbf{h}}^{t+1}, \mathbf{u}^t\right),\ \bar{\mathbf{h}}^{t+1} = \rho^{h\to u}\left(\mathbf{H}^{t+1}\right).$$

Here $r_k$ is the receiver node and $s_k$ is the sender node of edge $k$. $\mathbf{E}^{t+1}$ and $\mathbf{H}^{t+1}$ are the matrices of stacked edge vectors and node vectors at time step t+1, respectively. $\mathbf{E}_v^{t+1}$ collects edge vectors with receiver node $v$. $\mathbf{u}$ is the global attribute for graph representation. The $\varphi$ and $\rho$ functions can have various settings and the $\rho$ functions must be invariant to input orders and should take variable lengths of arguments.

### 3.2. Propagation modules - recurrent operator

Recurrent methods are pioneers in this research line. The major difference between recurrent operators and convolution operators is that layers in convolution operators use different weights while layers in recurrent operators share same weights. Early methods based on recursive neural networks focus on dealing with directed acyclic graphs (Sperduti and Starita, 1997; Frasconi et al., 1998; Micheli et al., 2004; Hammer et al., 2004). Later, the concept of graph neural network (GNN) was first proposed in (Scarselli et al., 2009; Gori et al., 2005), which extended existing neural networks to process more graph types. We name the model as GNN in this paper to distinguish it with the general name. We first introduce GNN and its later variants which require convergence of the hidden states and then we talk about methods based on the gate mechanism.

**Table 2**
Different variants of recurrent operators.

| Variant | Aggregator | Updater |
|---|---|---|
| GGNN | $\mathbf{h}^t_{\mathcal{N}_v} = \sum_{k\in\mathcal{N}_v} \mathbf{h}_k^{t-1} + \mathbf{b}$ | $\mathbf{z}_v^t = \sigma(\mathbf{W}^z\mathbf{h}^t_{\mathcal{N}_v} + \mathbf{U}^z\mathbf{h}_v^{t-1})$<br>$\mathbf{r}_v^t = \sigma(\mathbf{W}^r\mathbf{h}^t_{\mathcal{N}_v} + \mathbf{U}^r\mathbf{h}_v^{t-1})$<br>$\tilde{\mathbf{h}}_v^t = \tanh(\mathbf{W}\mathbf{h}^t_{\mathcal{N}_v} + \mathbf{U}(\mathbf{r}_v^t \odot \mathbf{h}_v^{t-1}))$<br>$\mathbf{h}_v^t = (1 - \mathbf{z}_v^t) \odot \mathbf{h}_v^{t-1} + \mathbf{z}_v^t \odot \tilde{\mathbf{h}}_v^t$ |
| Tree LSTM (Child sum) | $\mathbf{h}^{ti}_{\mathcal{N}_v} = \sum_{k\in\mathcal{N}_v} \mathbf{U}^i\mathbf{h}_k^{t-1}$<br>$\mathbf{h}^{tf}_{\mathcal{N}_v k} = \mathbf{U}^f\mathbf{h}_k^{t-1}$<br>$\mathbf{h}^{to}_{\mathcal{N}_v} = \sum_{k\in\mathcal{N}_v} \mathbf{U}^o\mathbf{h}_k^{t-1}$<br>$\mathbf{h}^{tu}_{\mathcal{N}_v} = \sum_{k\in\mathcal{N}_v} \mathbf{U}^u\mathbf{h}_k^{t-1}$ | $\mathbf{i}_v^t = \sigma(\mathbf{W}^i\mathbf{x}_v^t + \mathbf{h}^{ti}_{\mathcal{N}_v} + \mathbf{b}^i)$<br>$\mathbf{f}_{vk}^t = \sigma(\mathbf{W}^f\mathbf{x}_v^t + \mathbf{h}^{tf}_{\mathcal{N}_v k} + \mathbf{b}^f)$<br>$\mathbf{o}_v^t = \sigma(\mathbf{W}^o\mathbf{x}_v^t + \mathbf{h}^{to}_{\mathcal{N}_v} + \mathbf{b}^o)$<br>$\mathbf{u}_v^t = \tanh(\mathbf{W}^u\mathbf{x}_v^t + \mathbf{h}^{tu}_{\mathcal{N}_v} + \mathbf{b}^u)$<br>$\mathbf{c}_v^t = \mathbf{i}_v^t \odot \mathbf{u}_v^t + \sum_{k\in\mathcal{N}_v} \mathbf{f}_{vk}^t \odot \mathbf{c}_k^{t-1}$<br>$\mathbf{h}_v^t = \mathbf{o}_v^t \odot \tanh(\mathbf{c}_v^t)$ |
| Tree LSTM (N-ary) | $\mathbf{h}^{ti}_{\mathcal{N}_v} = \sum_{l=1}^{K} \mathbf{U}_l^i\mathbf{h}_{vl}^{t-1}$<br>$\mathbf{h}^{tf}_{\mathcal{N}_v k} = \sum_{l=1}^{K} \mathbf{U}_{kl}^f\mathbf{h}_{vl}^{t-1}$<br>$\mathbf{h}^{to}_{\mathcal{N}_v} = \sum_{l=1}^{K} \mathbf{U}_l^o\mathbf{h}_{vl}^{t-1}$<br>$\mathbf{h}^{tu}_{\mathcal{N}_v} = \sum_{l=1}^{K} \mathbf{U}_l^u\mathbf{h}_{vl}^{t-1}$ | |
| Graph LSTM in (Peng et al., 2017) | $\mathbf{h}^{ti}_{\mathcal{N}_v} = \sum_{k\in\mathcal{N}_v} \mathbf{U}^i_{m(v,k)}\mathbf{h}_k^{t-1}$<br>$\mathbf{h}^{tf}_{\mathcal{N}_v k} = \mathbf{U}^f_{m(v,k)}\mathbf{h}_k^{t-1}$<br>$\mathbf{h}^{to}_{\mathcal{N}_v} = \sum_{k\in\mathcal{N}_v} \mathbf{U}^o_{m(v,k)}\mathbf{h}_k^{t-1}$<br>$\mathbf{h}^{tu}_{\mathcal{N}_v} = \sum_{k\in\mathcal{N}_v} \mathbf{U}^u_{m(v,k)}\mathbf{h}_k^{t-1}$ | |

### 3.2.1. Convergence-based methods

In a graph, each node is naturally defined by its features and the related nodes. The target of GNN is to learn a state embedding $\mathbf{h}_v \in \mathbb{R}^s$ which contains the information of the neighborhood and itself for each node. The state embedding $\mathbf{h}_v$ is an $s$-dimension vector of node $v$ and can be used to produce an output $\mathbf{o}_v$ such as the distribution of the predicted node label. Then the computation steps of $\mathbf{h}_v$ and $\mathbf{o}_v$ are defined as:

$$\begin{aligned}\mathbf{h}_v &= f\left(\mathbf{x}_v, \mathbf{x}_{co[v]}, \mathbf{h}_{\mathcal{N}_v}, \mathbf{x}_{\mathcal{N}_v}\right),\\ \mathbf{o}_v &= g(\mathbf{h}_v, \mathbf{x}_v),\end{aligned} \tag{19}$$

where $\mathbf{x}_v, \mathbf{x}_{co[v]}, \mathbf{h}_{\mathcal{N}_v}, \mathbf{x}_{\mathcal{N}_v}$ are the features of $v$, the features of its edges, the states and the features of the nodes in the neighborhood of $v$, respectively. $f$ here is a parametric function called the *local transition function*. It is shared among all nodes and updates the node state according to the input neighborhood. $g$ is the *local output function* that describes how the output is produced. Note that both $f$ and $g$ can be interpreted as the feedforward neural networks.

Let $\mathbf{H}$, $\mathbf{O}$, $\mathbf{X}$, and $\mathbf{X}_N$ be the matrices constructed by stacking all the states, all the outputs, all the features, and all the node features, respectively. Then we have a compact form as:

$$\begin{aligned}\mathbf{H} &= F(\mathbf{H}, \mathbf{X}),\\ \mathbf{O} &= G(\mathbf{H}, \mathbf{X}_N),\end{aligned} \tag{20}$$

where $F$, the *global transition function*, and $G$, the *global output function* are stacked versions of $f$ and $g$ for all nodes in a graph, respectively. The value of $\mathbf{H}$ is the fixed point of Eq. (20) and is uniquely defined with the assumption that $F$ is a contraction map.

With the suggestion of Banach's fixed point theorem (Khamsi and Kirk, 2011), GNN uses the following classic iterative scheme to compute the state:

$$\mathbf{H}^{t+1} = F(\mathbf{H}^t, \mathbf{X}), \tag{21}$$

where $\mathbf{H}^t$ denotes the $t$-th iteration of $\mathbf{H}$. The dynamical system Eq. (21) converges exponentially fast to the solution for any initial value.

Though experimental results have shown that GNN is a powerful architecture for modeling structural data, there are still several limitations:

- GNN requires $f$ to be a contraction map which limits the model's ability. And it is inefficient to update the hidden states of nodes iteratively towards the fixed point.
- It is unsuitable to use the fixed points if we focus on the representation of nodes instead of graphs because the distribution of representation in the fixed point will be much smoother in value and less informative for distinguishing each node.

*GraphESN.* Graph echo state network (GraphESN) (Gallicchio and Micheli, 2010) generalizes the echo state network (ESN) (Jaeger, 2001) on graphs. It uses a fixed contractive encoding function, and only trains a readout function. The convergence is ensured by the contractivity of reservoir dynamics. As a consequence, GraphESN is more efficient than GNN.

*SSE.* Stochastic Steady-state Embedding (SSE) (Dai et al., 2018a) is also proposed to improve the efficiency of GNN. SSE proposes a learning framework which contains two steps. Embeddings of each node are updated by a parameterized operator in the update step and these embeddings are projected to the steady state constraint space to meet the steady-state conditions.

*LP-GNN.* Lagrangian Propagation GNN (LP-GNN) (Tiezzi et al., 2020) formalizes the learning task as a constraint optimization problem in the Lagrangian framework and avoids the iterative computations for the fixed point. The convergence procedure is implicitly expressed by a constraint satisfaction mechanism.

### 3.2.2. Gate-based methods

There are several works attempting to use the gate mechanism like GRU (Cho et al., 2014) or LSTM (Hochreiter and Schmidhuber, 1997) in the propagation step to diminish the computational limitations in GNN and improve the long-term propagation of information across the graph structure. They run a fixed number of training steps without the guarantee of convergence.

*GGNN.* The gated graph neural network (GGNN) (Li et al., 2016) is proposed to release the limitations of GNN. It releases the requirement of function $f$ to be a contraction map and uses the Gate Recurrent Units (GRU) in the propagation step. It also uses back-propagation through time (BPTT) to compute gradients. The computation step of GGNN can be found in Table 2.

The node $v$ first aggregates messages from its neighbors. Then the GRU-like update functions incorporate information from the other nodes and from the previous timestep to update each node's hidden state. $\mathbf{h}_{\mathcal{N}_v}$ gathers the neighborhood information of node $v$, while $\mathbf{z}$ and $\mathbf{r}$ are the update and reset gates.

LSTMs are also used in a similar way as GRU through the propagation process based on a tree or a graph.

*Tree LSTM.* Tai et al. (2015) propose two extensions on the tree structure to the basic LSTM architecture: the *Child-Sum Tree-LSTM* and the *N-ary Tree-LSTM*. They are also extensions to the recursive neural network based models as we mentioned before. Tree is a special case of graph and each node in Tree-LSTM aggregates information from its children. Instead of a single forget gate in traditional LSTM, the Tree-LSTM unit for node $v$ contains one forget gate $\mathbf{f}_{vk}$ for each child $k$. The computation step of the Child-Sum Tree-LSTM is displayed in Table 2. $\mathbf{i}_v^t$, $\mathbf{o}_v^t$, and $\mathbf{c}_v^t$ are the input gate, output gate and memory cell respectively. $\mathbf{x}_v^t$ is the input vector at time $t$. The $N$-ary Tree-LSTM is further designed for a special kind of tree where each node has at most $K$ children and the children are ordered. The equations for computing $\mathbf{h}_{\mathcal{N}_v}^{ti}$, $\mathbf{h}_{\mathcal{N}_v k}^{tf}$, $\mathbf{h}_{\mathcal{N}_v}^{to}$, $\mathbf{h}_{\mathcal{N}_v}^{tu}$ in Table 2 introduce separate parameters for each child $k$. These parameters allow the model to learn more fine-grained representations conditioning on the states of a unit's children than the Child-Sum Tree-LSTM.

*Graph LSTM.* The two types of Tree-LSTMs can be easily adapted to the graph. The graph-structured LSTM in (Zayats and Ostendorf, 2018) is an example of the $N$-ary Tree-LSTM applied to the graph. However, it is a simplified version since each node in the graph has at most 2 incoming edges (from its parent and sibling predecessor). Peng et al. (2017) propose another variant of the Graph LSTM based on the relation extraction task. The edges of graphs in (Peng et al., 2017) have various labels so that Peng et al. (2017) utilize different weight matrices to represent different labels. In Table 2, $m(v,k)$ denotes the edge label between node $v$ and $k$.

Liang et al. (2016) propose a Graph LSTM network to address the semantic object parsing task. It uses the confidence-driven scheme to adaptively select the starting node and determine the node updating sequence. It follows the same idea of generalizing the existing LSTMs into the graph-structured data but has a specific updating sequence while methods mentioned above are agnostic to the order of nodes.

*S-LSTM.* Zhang et al. (2018d) propose Sentence LSTM (S-LSTM) for improving text encoding. It converts text into a graph and utilizes the Graph LSTM to learn the representation. The S-LSTM shows strong representation power in many NLP problems.

### 3.3. *Propagation modules - skip connection*

Many applications unroll or stack the graph neural network layer aiming to achieve better results as more layers (i.e $k$ layers) make each node aggregate more information from neighbors $k$ hops away. However, it has been observed in many experiments that deeper models could not improve the performance and deeper models could even perform worse. This is mainly because more layers could also propagate the noisy information from an exponentially increasing number of expanded neighborhood members. It also causes the over smoothing problem because nodes tend to have similar representations after the aggregation operation when models go deeper. So that many methods try to add "skip connections" to make GNN models deeper. In this subsection we introduce three kinds of instantiations of skip connections.

*Highway GCN.* Rahimi et al. (2018) propose a Highway GCN which uses layer-wise gates similar to highway networks (Zilly et al., 2016). The output of a layer is summed with its input with gating weights:

$$\begin{aligned} \mathbf{T}(\mathbf{h}^t) &= \sigma(\mathbf{W}_t\mathbf{h}^t + \mathbf{b}_t), \\ \mathbf{h}^{t+1} &= \mathbf{h}^{t+1} \odot \mathbf{T}(\mathbf{h}^t) + \mathbf{h}^t \odot (1 - \mathbf{T}(\mathbf{h}^t)). \end{aligned} \tag{22}$$

By adding the highway gates, the performance peaks at 4 layers in a specific problem discussed in (Rahimi et al., 2018). The column network (CLN) (Pham et al., 2017) also utilizes the highway network. But it has different functions to compute the gating weights.

*JKN.* Xu et al. (2018) study properties and limitations of neighborhood aggregation schemes. They propose the jump knowledge network (JKN) which could learn adaptive and structure-aware representations. JKN selects from all of the intermediate representations (which "jump" to the last layer) for each node at the last layer, which makes the model adapt the effective neighborhood size for each node as needed. Xu et al. (2018) use three approaches of concatenation, max-pooling and LSTM-attention in the experiments to aggregate information. The JKN performs well on the experiments in social, bioinformatics and citation networks. It can also be combined with models like GCN, GraphSAGE and GAT to improve their performance.

*DeepGCNs.* Li et al. (2019a) borrow ideas from ResNet (He et al., 2016a, 2016b) and DenseNet (Huang et al., 2017). ResGCN and DenseGCN are proposed by incorporating residual connections and dense connections to solve the problems of vanishing gradient and over smoothing. In detail, the hidden state of a node in ResGCN and DenseGCN can be computed as:

$$\begin{aligned} \mathbf{h}_{\text{Res}}^{t+1} &= \mathbf{h}^{t+1} + \mathbf{h}^t, \\ \mathbf{h}_{\text{Dense}}^{t+1} &= \Vert_{i=0}^{t+1}\mathbf{h}^i. \end{aligned} \tag{23}$$

The experiments of DeepGCNs are conducted on the point cloud semantic segmentation task and the best results are achieved with a 56-layer model.

### 3.4. *Sampling modules*

GNN models aggregate messages for each node from its neighborhood in the previous layer. Intuitively, if we track back multiple GNN layers, the size of supporting neighbors will grow exponentially with the depth. To alleviate this "neighbor explosion" issue, an efficient and efficacious way is sampling. Besides, when we deal with large graphs, we cannot always store and process all neighborhood information for each node, thus the sampling module is needed to conduct the propagation. In this section, we introduce three kinds of graph sampling modules: node sampling, layer sampling, and subgraph sampling.

#### 3.4.1. *Node sampling*

A straightforward way to reduce the size of neighboring nodes would be selecting a subset from each node's neighborhood. GraphSAGE (Hamilton et al., 2017a) samples a fixed small number of neighbors, ensuring a 2 to 50 neighborhood size for each node. To reduce sampling variance, Chen et al. (2018a) introduce a control-variate based stochastic approximation algorithm for GCN by utilizing the historical activations of nodes as a control variate. This method limits the receptive field in the 1-hop neighborhood, and uses the historical hidden state as an affordable approximation.

PinSage (Ying et al., 2018a) proposes importance-based sampling method. By simulating random walks starting from target nodes, this approach chooses the top T nodes with the highest normalized visit counts.

#### 3.4.2. *Layer sampling*

Instead of sampling neighbors for each node, layer sampling retains a small set of nodes for aggregation in each layer to control the expansion factor. FastGCN (Chen et al., 2018b) directly samples the receptive field for each layer. It uses importance sampling, where the important nodes are more likely to be sampled.

In contrast to fixed sampling methods above, Huang et al. (2018) introduce a parameterized and trainable sampler to perform layer-wise sampling conditioned on the former layer. Furthermore, this adaptive sampler could optimize the sampling importance and reduce variance simultaneously. LADIES (Zou et al., 2019) intends to alleviate the sparsity issue in layer-wise sampling by generating samples from the union of neighbors of the nodes.

#### 3.4.3. *Subgraph sampling*

Rather than sampling nodes and edges which builds upon the full graph, a fundamentally different way is to sample multiple subgraphs and restrict the neighborhood search within these subgraphs. ClusterGCN (Chiang et al., 2019) samples subgraphs by graph clustering algorithms, while GraphSAINT (Zeng et al., 2020) directly samples nodes or edges to generate a subgraph.

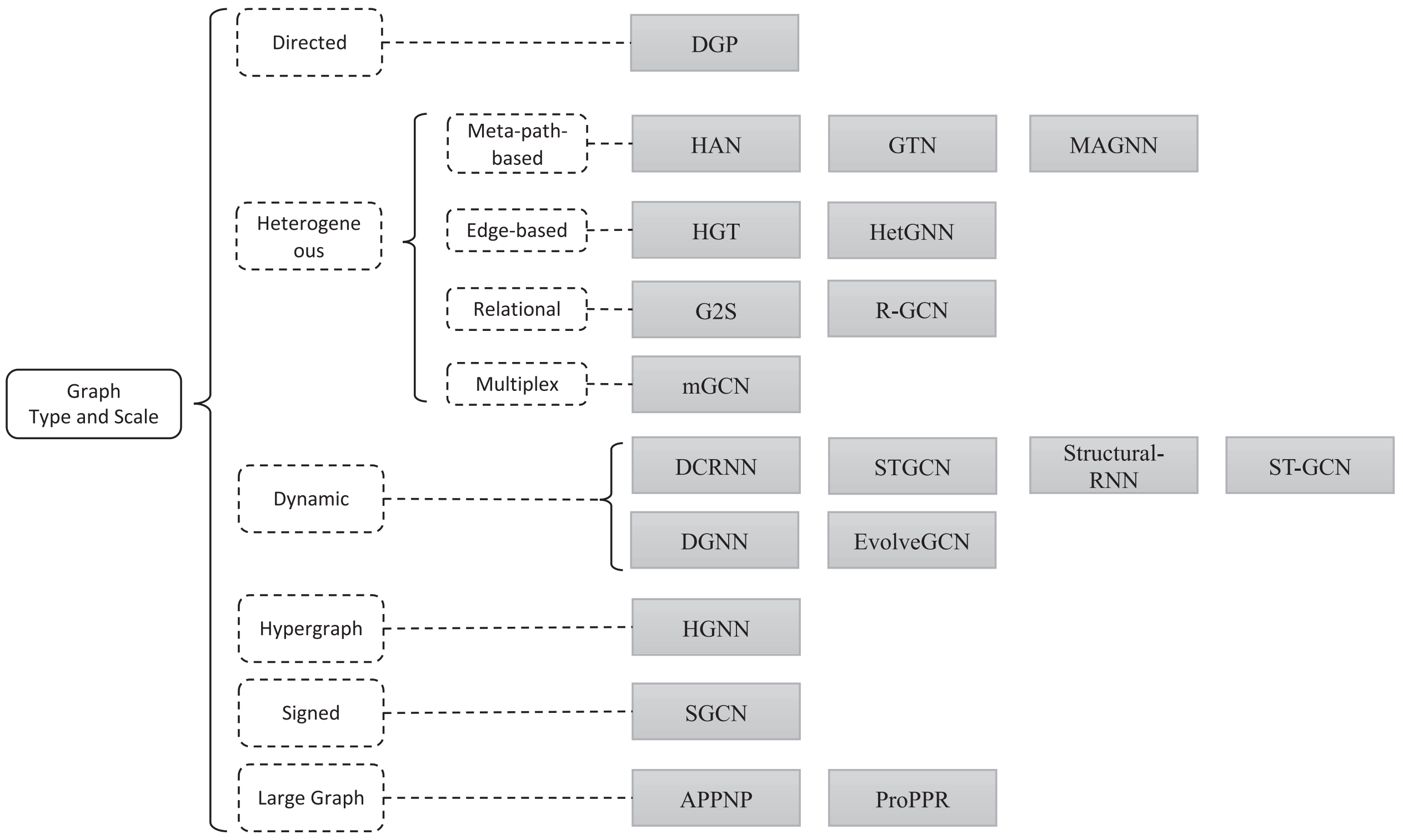

**Fig. 4.** An overview of variants considering graph type and scale.

### *3.5. Pooling modules*

In the area of computer vision, a convolutional layer is usually followed by a pooling layer to get more general features. Complicated and large-scale graphs usually carry rich hierarchical structures which are of great importance for node-level and graph-level classification tasks. Similar to these pooling layers, a lot of work focuses on designing hierarchical pooling layers on graphs. In this section, we introduce two kinds of pooling modules: direct pooling modules and hierarchical pooling modules.

#### *3.5.1. Direct pooling modules*

Direct pooling modules learn graph-level representations directly from nodes with different node selection strategies. These modules are also called readout functions in some variants.

*Simple Node Pooling.* Simple node pooling methods are used by several models. In these models, node-wise max/mean/sum/attention operations are applied on node features to get a global graph representation.

*Set2set.* MPNN uses the Set2set method (Vinyals et al., 2015a) as the readout function to get graph representations. Set2set is designed to deal with the unordered set $T = \{(\mathbf{h}_v^T, \mathbf{x}_v)\}$ and uses a LSTM-based method to produce an order invariant representation after a predifined number of steps.

*SortPooling.* SortPooling (Zhang et al., 2018e) first sorts the node embeddings according to the structural roles of the nodes and then the sorted embeddings are fed into CNNs to get the representation.

#### *3.5.2. Hierarchical pooling modules*

The methods mentioned before directly learn graph representations from nodes and they do not investigate the hierarchical property of the graph structure. Next we will talk about methods that follow a hierarchical pooling pattern and learn graph representations by layers.

*Graph Coarsening.* Early methods are usually based on graph coarsening algorithms. Spectral clustering algorithms are firstly used but they are inefficient because of the eigendecomposition step. Graclus (Dhillon et al., 2007) provides a faster way to cluster nodes and it is applied as a pooling module. For example, ChebNet and MoNet use Graclus to merge node pairs and further add additional nodes to make sure the pooling procedure forms a balanced binary tree.

*ECC.* Edge-Conditioned Convolution (ECC) (Simonovsky and Komodakis, 2017) designs its pooling module with recursively downsampling operation. The downsampling method is based on splitting the graph into two components by the sign of the largest eigenvector of the Laplacian.

*DiffPool.* DiffPool (Ying et al., 2018b) uses a learnable hierarchical clustering module by training an assignment matrix $\mathbf{S}^t$ in each layer:

$$\begin{aligned} \mathbf{S}^t &= \text{softmax}\left(\text{GNN}_{t,pool}(\mathbf{A}^t, \mathbf{H}^t)\right), \\ \mathbf{A}^{t+1} &= (\mathbf{S}^t)^T \mathbf{A}^t \mathbf{S}^t, \end{aligned} \tag{24}$$

where $\mathbf{H}^t$ is the node feature matrix and $\mathbf{A}^t$ is coarsened adjacency matrix of layer $t$. $\mathbf{S}^t$ denotes the probabilities that a node in layer $t$ can be assigned to a coarser node in layer $t+1$.

*gPool.* gPool (Gao and Ji, 2019) uses a project vector to learn projection scores for each node and select nodes with top-k scores. Compared to DiffPool, it uses a vector instead of a matrix at each layer, thus it reduces the storage complexity. But the projection procedure does not consider the graph structure.

*EigenPooling.* EigenPooling (Ma et al., 2019a) is designed to use the node features and local structure jointly. It uses the local graph Fourier transform to extract subgraph information and suffers from the inefficiency of graph eigendecomposition.

*SAGPool.* SAGPool (Lee et al., 2019) is also proposed to use features and topology jointly to learn graph representations. It uses a self-attention based method with a reasonable time and space complexity.

## 4. Variants considering graph type and scale

In the above sections, we assume the graph to be the simplest format. However, many graphs in the real world are complex. In this subsection, we will introduce the approaches which attempt to address the challenges of complex graph types. An overview of these variants is shown in Fig. 4.

### 4.1. Directed graphs

The first type is the directed graphs. Directed edges usually contain more information than undirected edges. For example, in a knowledge graph where a head entity is the parent class of a tail entity, the edge direction offers information about the partial order. Instead of simply adopting an asymmetric adjacency matrix in the convolution operator, we can model the forward and reverse directions of an edge differently. DGP (Kampffmeyer et al., 2019) uses two kinds of weight matrices $\mathbf{W}_p$ and $\mathbf{W}_c$ for the convolution in forward and reverse directions.

### 4.2. Heterogeneous graphs

The second variant of graphs is heterogeneous graphs, where the nodes and edges are multi-typed or multi-modal. More specifically, in a heterogeneous graph $\{V, E, \varphi, \psi\}$, each node $v_i$ is associated with a type $\varphi(v_i)$ and each edge $e_j$ with a type $\psi(e_j)$.

#### 4.2.1. Meta-path-based methods

Most approaches toward this graph type utilize the concept of meta-path. Meta-path is a path scheme which determines the type of node in each position of the path, e.g. $\varphi_1 \xrightarrow{\psi_1} \varphi_2 \xrightarrow{\psi_2} \varphi_3 \cdots \xrightarrow{\psi_L} \varphi_{L+1}$, where $L$ is the length of the meta-path. In the training process, the meta-paths are instantiated as node sequences. By connecting the two end nodes of a meta-path instances, the meta-path captures the similarity of two nodes which may not be directly connected. Consequently, one heterogeneous graph can be reduced to several homogeneous graphs, on which graph learning algorithms can be applied. In early work, meta-path based similarity search is investigated (Sun et al., 2011). Recently, more GNN models which utilize the meta-path are proposed. HAN (Wang et al., 2019a) first performs graph attention on the meta-path-based neighbors under each meta-path and then uses a semantic attention over output embeddings of nodes under all meta-path schemes to generate the final representation of nodes. MAGNN (Fu et al., 2020) proposes to take the intermediate nodes in a meta-path into consideration. It first aggregates the information along the meta-path using a neural module and then performs attention over different meta-path instances associated with a node and finally performs attention over different meta-path schemes. GTN (Yun et al., 2019) proposes a novel graph transformer layer which identifies new connections between unconnected nodes while learning representations of nodes. The learned new connections can connect nodes which are serveral hops away from each other but are closely related, which function as the meta-paths.

#### 4.2.2. Edge-based methods

There are also works which don't utilize meta-paths. These works typically use different functions in terms of sampling, aggregation, etc. for different kinds of neighbors and edges. HetGNN (Zhang et al., 2019b) addresses the challenge by directly treating neighbors of different types differently in sampling, feature encoding and aggregation steps. HGT (Hu et al., 2020a) defines the meta-relation to be the type of two neighboring nodes and their link $\langle \varphi(v_i), \psi(e_{ij}), \varphi(v_j) \rangle$. It assigns different attention weight matrices to different meta-relations, empowering the model to take type information into consideration.

#### 4.2.3. Methods for relational graphs

The edge of some graphs may contain more information than the type, or the quantity of types may be too large, exerting difficulties to applying the meta-path or meta-relation based methods. We refer to this kind of graphs as relational graphs (Schlichtkrull et al., 2018), To handle the relational graphs, G2S (Beck et al., 2018) converts the original graph to a bipartite graph where the original edges also become nodes and one original edge is split into two new edges which means there are two new edges between the edge node and begin/end nodes. After this transformation, it uses a Gated Graph Neural Network followed by a Recurrent Neural Network to convert graphs with edge information into sentences. The aggregation function of GGNN takes both the hidden representations of nodes and the relations as the input. As another approach, R-GCN (Schlichtkrull et al., 2018) doesn't require to convert the original graph format. It assigns different weight matrices for the propagation on different kinds of edges. However, When the number of relations is very large, the number of parameters in the model explodes. Therefore, it introduces two kinds of regularizations to reduce the number of parameters for modeling amounts of relations: *basis-* and *block-diagonal*-decomposition. With the basis decomposition, each $\mathbf{W}_r$ is defined as follows:

$$\mathbf{W}_r = \sum_{b=1}^{B} a_{rb} \mathbf{V}_b. \tag{25}$$

Here each $\mathbf{W}_r$ is a linear combination of basis transformations $\mathbf{V}_b \in \mathbb{R}^{d_{in} \times d_{out}}$ with coefficients $a_{rb}$. In the block-diagonal decomposition, R-GCN defines each $\mathbf{W}_r$ through the direct sum over a set of low-dimensional matrices, which need more parameters than the first one.

#### 4.2.4. Methods for multiplex graphs

In more complex scenarios, a pair of nodes in a graph can be associated with multiple edges of different types. By viewing under different types of edges, the graph can form multiple layers, in which each layer represents one type of relation. Therefore, multiplex graph can also be referred to as multi-view graph (multi-dimensional graph). For example, in YouTube, there can be three different relations between two users: sharing, subscription, comment. Edge types are not assumed independent with each other, therefore simply splitting the graph into subgraphs with one type of edges might not be an optimal solution. mGCN (Ma et al., 2019b) introduces general representations and dimension-specific representations for nodes in each layer of GNN. The dimension-specific representations are projected from general representations using

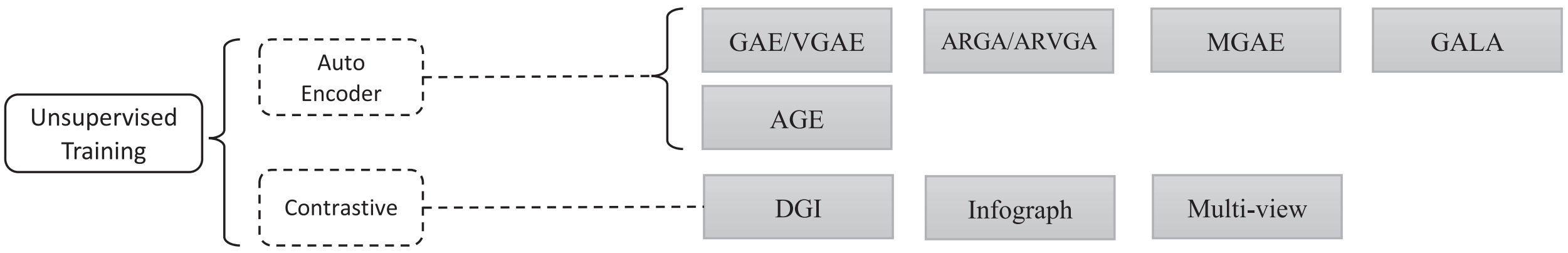

**Fig. 5.** An overview of methods with unsupervised loss.

different projection matrices and then aggregated to form the next layer's general representations.

### 4.3. Dynamic graphs

Another variant of graphs is dynamic graphs, in which the graph structure, e.g. the existence of edges and nodes, keeps changing over time. To model the graph structured data together with the time series data, DCRNN (Li et al., 2018b) and STGCN (Yu et al., 2018) first collect spatial information by GNNs, then feed the outputs into a sequence model like sequence-to-sequence models or RNNs. Differently, Structural-RNN (Jain et al., 2016) and ST-GCN (Yan et al., 2018) collect spatial and temporal messages at the same time. They extend static graph structure with temporal connections so they can apply traditional GNNs on the extended graphs. Similarly, DGNN (Manessi et al., 2020) feeds the output embeddings of each node from the GCN into separate LSTMs. The weights of LSTMs are shared between each node. On the other hand, EvolveGCN (Pareja et al., 2020) argues that directly modeling dynamics of the node representation will hamper the model's performance on graphs where node set keeps changing. Therefore, instead of treating node features as the input to RNN, it feeds the weights of the GCN into the RNN to capture the intrinsic dynamics of the graph interactions. Recently, a survey (Huang et al., 2020) classifies the dynamic networks into several categories based on the link duration, and groups the existing models into these categories according to their specialization. It also establishes a general framework for models of dynamic graphs and fits existing models into the general framework.

### 4.4. Other graph types

For other variants of graphs, such as hypergraphs and signed graphs, there are also some models proposed to address the challenges.

#### 4.4.1. Hypergraphs

A hypergraph can be denoted by $G = (V, E, W_e)$, where an edge $e \in E$ connects two or more vertices and is assigned a weight $w \in W_e$. The adjacency matrix of a hypergraph can be represented in a $|V| \times |E|$ matrix $\mathbf{L}$:

$$L_{v,e} = \left\{ \begin{matrix} 1, & \text{if} \quad v \in e \\ 0, & \text{if} \quad v \notin e \end{matrix} \right\}. \tag{26}$$

HGNN (Feng et al., 2019) proposes hypergraph convolution to process these high order interaction between nodes:

$$\mathbf{H} = \mathbf{D}_v^{-\frac{1}{2}}\mathbf{L}\mathbf{W}_e\mathbf{D}_e^{-1}\mathbf{L}^T\mathbf{D}_v^{-\frac{1}{2}}\mathbf{X}\mathbf{W}, \tag{27}$$

where the $\mathbf{D}_v, \mathbf{W}_e, \mathbf{D}_e, \mathbf{X}$ are the node degree matrix, edge weight matrix, edge degree matrix and node feature matrix respectively. $\mathbf{W}$ is the learnable parameters. This formula is derived by approximating the hypergraph Laplacian using truncated Chebyshev polynomials.

#### 4.4.2. Signed graphs

Signed graphs are the graphs with signed edges, i.e. an edge can be either positive or negative. Instead of simply treating the negative edges as the absent edges or another type of edges, SGCN (Derr et al., 2018) utilizes balance theory to capture the interactions between positive edges and negative edges. Intuitively, balance theory suggests that the friend (positive edge) of my friend is also my friend and the enemy (negative edge) of my enemy is my friend. Therefore it provides theoretical foundation for SGCN to model the interactions between positive edges and negative edges.

### 4.5. Large graphs

As we mentioned in Section 3.4, sampling operators are usually used to process large-scale graphs. Besides sampling techniques, there are also other methods for the scaling problem. Leveraging approximate personalized PageRank, methods proposed by Klicpera et al. (2019) and Bojchevski et al. (2020) avoid calculating high-order propagation matrices. Rossi et al. (2020) propose a method to precompute graph convolutional filters of different sizes for efficient training and inference. PageRank-based models squeeze multiple GCN layers into one single propagation layer to mitigate the "neighbor explosion" issue, hence are highly scalable and efficient.

## 5. Variants for different training settings

In this section, we introduce variants for different training settings. For supervised and semi-supervised settings, labels are provided so that loss functions are easy to design for these labeled samples. For unsupervised settings, there are no labeled samples so that loss functions should depend on the information provided by the graph itself, such as input features or the graph topology. In this section, we mainly introduce variants for unsupervised training, which are usually based on the ideas of auto-encoders or contrastive learning. An overview of the methods we mention is shown in Fig. 5.

### 5.1. Graph auto-encoders

For unsupervised graph representation learning, there has been a trend to extend auto-encoder (AE) to graph domains.

Graph Auto-Encoder (GAE) (Kipf and Welling, 2016) first uses GCNs to encode nodes in the graph. Then it uses a simple decoder to reconstruct the adjacency matrix and compute the loss from the similarity between the original adjacency matrix and the reconstructed matrix:

$$\begin{aligned} \mathbf{H} &= \text{GCN}(\mathbf{X}, \mathbf{A}), \\ \tilde{\mathbf{A}} &= \rho(\mathbf{H}\mathbf{H}^T). \end{aligned} \tag{28}$$

Kipf and Welling (2016) also train the GAE model in a variational manner and the model is named as the variational graph auto-encoder (VGAE).

Adversarially Regularized Graph Auto-encoder (ARGA) (Pan et al., 2018) employs generative adversarial networks (GANs) to regularize a GCN-based graph auto-encoder, which could learn more robust node representations.

Instead of recovering the adjacency matrix, Wang et al. (2017), Park et al. (2019) try to reconstruct the feature matrix. MGAE (Wang et al., 2017) utilizes marginalized denoising auto-encoder to get robust node representation. To build a symmetric graph auto-encoder, GALA (Park et al., 2019) proposes Laplacian sharpening, the inverse operation of Laplacian smoothing, to decode hidden states. This mechanism alleviates the oversmoothing issue in GNN training.

Different from above, AGE (Cui et al., 2020) states that the recovering losses are not compatible with downstream tasks. Therefore, they apply adaptive learning for the measurement of pairwise node similarity and achieve state-of-the-art performance on node clustering and link prediction.

### 5.2. Contrastive learning

Besides graph auto-encoders, contrastive learning paves another way for unsupervised graph representation learning. Deep Graph Infomax (DGI) (Velickovic et al., 2019) maximizes mutual information between node representations and graph representations. Infograph (Sun et al., 2020) aims to learn graph representations by mutual information maximization between graph-level representations and the substructure-level representations of different scales including nodes, edges and triangles. Multi-view (Hassani and Khasahmadi, 2020) contrasts representations from first-order adjacency matrix and graph diffusion, achieves state-of-the-art performances on multiple graph learning tasks.

## 6. A design example of GNN

In this section, we give an existing GNN model to illustrated the design process. Taking the task of heterogeneous graph pretraining as an example, we use GPT-GNN (Hu et al., 2020b) as the model to illustrate the design process.

1. *Find graph structure.* The paper focuses on applications on the academic knowledge graph and the recommendation system. In the academic knowledge graph, the graph structure is explicit. In recommendation systems, users, items and reviews can be regarded as nodes and the interactions among them can be regarded as edges, so the graph structure is also easy to construct.
2. *Specify graph type and scale.* The tasks focus on heterogeneous graphs, so that types of nodes and edges should be considered and incorporated in the final model. As the academic graph and the recommendation graph contain millions of nodes, so that the model should further consider the efficiency problem. In conclusion, the model should focus on large-scale heterogeneous graphs.
3. *Design loss function.* As downstream tasks in (Hu et al., 2020b) are all node-level tasks (e.g. Paper-Field prediction in the academic graph), so that the model should learn node representations in the pretraining step. In the pretraining step, no labeled data is available, so that a self-supervised graph generation task is designed to learn node embeddings. In the finetuning step, the model is finetuned based on the training data of each task, so that the supervised loss of each task is applied.
4. *Build model using computational modules.* Finally the model is built with computational modules. For the propagation module, the authors use a convolution operator HGT (Hu et al., 2020a) that we mentioned before. HGT incorporates the types of nodes and edges into the propagation step of the model and the skip connection is also added in the architecture. For the sampling module, a specially designed sampling method HGSampling (Hu et al., 2020a) is applied, which is a heterogeneous version of LADIES (Zou et al., 2019). As the model focuses on learning node representations, the pooling module is not needed. The HGT layer are stacked multiple layers to learn better node embeddings.

## 7. Analyses of GNNs

### 7.1. Theoretical aspect

In this section, we summarize the papers about the theoretic foundations and explanations of graph neural networks from various perspectives.

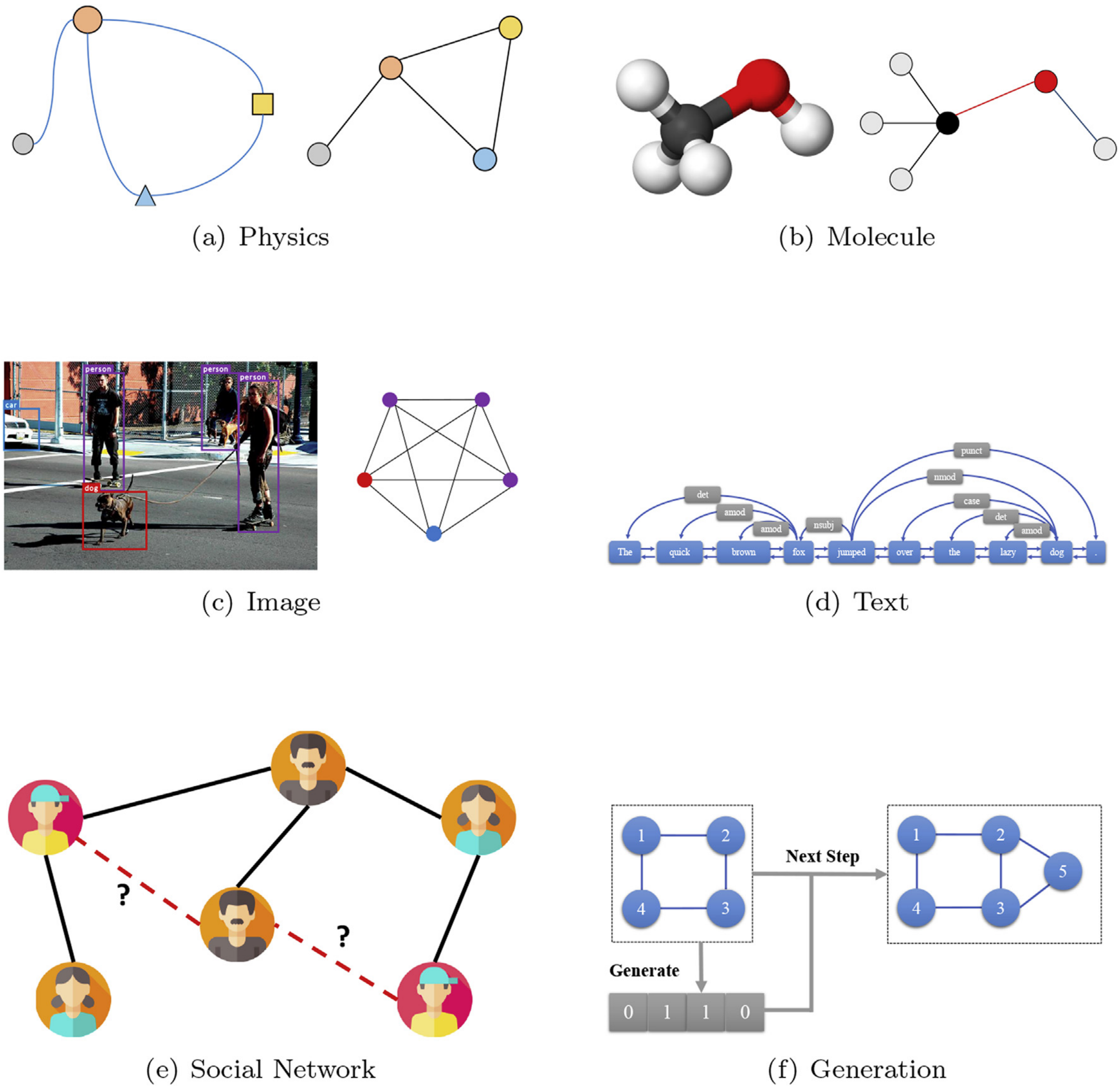

**Fig. 6.** Application scenarios. (Icons made by Freepik from Flaticon)

**Table 3**
Applications of graph neural networks.

| Area | Application | References |
|---|---|---|
| Graph Mining | Graph Matching | (Riba et al., 2018; Li et al., 2019b) |
| | Graph Clustering | (Zhang et al., 2019c; Ying et al., 2018b; Tsitsulin et al., 2020) |
| Physics | Physical Systems Modeling | (Battaglia et al., 2016; Sukhbaatar Ferguset al., 2016; Watters et al., 2017; Hoshen, 2017; Kipf et al., 2018; Sanchez et al., 2018) |
| Chemistry | Molecular Fingerprints | (Duvenaud et al., 2015; Kearnes et al., 2016) |
| | Chemical Reaction Prediction | Do et al. (2019) |
| Biology | Protein Interface Prediction | Fout et al. (2017) |
| | Side Effects Prediction | Zitnik et al. (2018) |
| | Disease Classification | Rhee et al. (2018) |
| Knowledge Graph | KB Completion | (Hamaguchi et al., 2017; Schlichtkrull et al., 2018; Shang et al., 2019) |
| | KG Alignment | (Wang et al., 2018b; Zhang et al., 2019d; Xu et al., 2019c) |
| Generation | Graph Generation | (Shchur et al., 2018b; Nowak et al., 2018; Ma et al., 2018; You et al., 2018a, 2018b; De Cao and Kipf, 2018; Li et al., 2018d; Shi et al., 2020; Liu et al., 2019; Grover et al., 2019) |
| Combinatorial Optimization | Combinatorial Optimization | (Khalil et al., 2017; Nowak et al., 2018; Li et al., 2018e; Kool et al., 2019; Bello et al., 2017; Vinyals et al., 2015b; Sutton and Barto, 2018; Dai et al., 2016; Gasse et al., 2019; Zheng et al., 2020a; Selsam et al., 2019; Sato et al., 2019) |
| Traffic Network | Traffic State Prediction | (Cui et al., 2018b; Yu et al., 2018; Zheng et al., 2020b; Guo et al., 2019) |
| Recommendation Systems | User-item Interaction Prediction | (van den Berg et al., 2017; Ying et al., 2018a) |
| | Social Recommendation | (Wu et al., 2019c; Fan et al., 2019) |
| Others (Structural) | Stock Market | (Matsunaga et al., 2019; Yang et al., 2019; Chen et al., 2018c; Li et al., 2020; Kim et al., 2019) |
| | Software Defined Networks | Rusek et al. (2019) |
| | AMR Graph to Text | (Song et al., 2018a; Beck et al., 2018) |
| Text | Text Classification | (Peng et al., 2018; Yao et al., 2019; Zhang et al., 2018d; Tai et al., 2015) |
| | Sequence Labeling | (Zhang et al., 2018d; Marcheggiani and Titov, 2017) |
| | Neural Machine Translation | (Bastings et al., 2017; Marcheggiani et al., 2018; Beck et al., 2018) |
| | Relation Extraction | (Miwa and Bansal, 2016; Peng et al., 2017; Song et al., 2018b; Zhang et al., 2018f) |
| | Event Extraction | (Nguyen and Grishman, 2018; Liu et al., 2018) |
| | Fact Verification | (Zhou et al., 2019; Liu et al., 2020; Zhong et al., 2020) |
| | Question Answering | (Song et al., 2018c; De Cao et al., 2019; Qiu et al., 2019; Tu et al., 2019; Ding et al., 2019) |
| | Relational Reasoning | (Santoro et al., 2017; Palm et al., 2018; Battaglia et al., 2016) |
| Image | Social Relationship Understanding | Wang et al. (2018c) |
| | Image Classification | (Garcia and Bruna, 2018; Wang et al., 2018d; Lee et al., 2018b; Kampffmeyer et al., 2019; Marino et al., 2017) |
| | Visual Question Answering | (Teney et al., 2017; Wang et al., 2018c; Narasimhan et al., 2018) |
| | Object Detection | (Hu et al., 2018; Gu et al., 2018) |
| | Interaction Detection | (Qi et al., 2018; Jain et al., 2016) |
| | Region Classification | Chen et al. (2018d) |
| | Semantic Segmentation | (Liang et al., 2016, 2017; Landrieu and Simonovsky, 2018; Wang et al., 2018e; Qi et al., 2017b) |
| Other (Non-structural) | Program Verification | (Allamanis et al., 2018; Li et al., 2016) |

#### 7.1.1. *Graph signal processing*

From the spectral perspective of view, GCNs perform convolution operation on the input features in the spectral domain, which follows graph signal processing in theory.

There exists several works analyzing GNNs from graph signal processing. Li et al. (2018c) first address the graph convolution in graph neural networks is actually Laplacian smoothing, which smooths the feature matrix so that nearby nodes have similar hidden representations. Laplacian smoothing reflects the homophily assumption that nearby nodes are supposed to be similar. The Laplacian matrix serves as a low-pass filter for the input features. SGC (Wu et al., 2019b) further removes the weight matrices and nonlinearties between layers, showing that the low-pass filter is the reason why GNNs work.

Following the idea of low-pass filtering, Zhang et al. (2019c), Cui et al. (2020), NT and Maehara (Nt and Maehara, 2019), Chen et al. (2020b) analyze different filters and provide new insights. To achieve low-pass filtering for all the eigenvalues, AGC (Zhang et al., 2019c) designs a graph filter $I - \frac{1}{2}L$ according to the frequency response function. AGE (Cui et al., 2020) further demonstrates that filter with $I - \frac{1}{\lambda_{max}}L$ could get better results, where $\lambda_{max}$ is the maximum eigenvalue of the Laplacian matrix. Despite linear filters, GraphHeat (Xu et al., 2019a) leverages heat kernels for better low-pass properties. NT and Maehara (Nt and Maehara, 2019) state that graph convolution is mainly a denoising process for input features, the model performances heavily depend on the amount of noises in the feature matrix. To alleviate the over-smoothing issue, Chen et al. (2020b) present two metrics for measuring the smoothness of node representations and the over-smoothness of GNN models. The authors conclude that the information-to-noise ratio is the key factor for over-smoothing.

#### 7.1.2. *Generalization*

The generalization ability of GNNs have also received attentions recently. Scarselli et al. (2018) prove the VC-dimensions for a limited class of GNNs. Garg et al. (2020) further give much tighter generalization bounds based on Rademacher bounds for neural networks.

Verma and Zhang (2019) analyze the stability and generalization properties of single-layer GNNs with different convolutional filters. The authors conclude that the stability of GNNs depends on the largest eigenvalue of the filters. Knyazev et al. (2019) focus on the generalization ability of attention mechanism in GNNs. Their conclusion shows that attention helps GNNs generalize to larger and noisy graphs.

#### 7.1.3. *Expressivity*

On the expressivity of GNNs, Xu et al. (2019b), Morris et al. (2019) show that GCNs and GraphSAGE are less discriminative than Weisfeiler-Leman (WL) test, an algorithm for graph isomorphism testing. Xu et al. (2019a) also propose GINs for more expressive GNNs. Going beyond WL test, Barceló et al. (2019) discuss if GNNs are expressible for $FOC_2$, a fragment of first order logic. The authors find that existing GNNs can hardly fit the logic. For learning graph topologic structures, Garg et al. (2020) prove that locally dependent GNN variants are not capable to learn global graph properties, including diameters, biggest/smallest cycles, or motifs.

Loukas (2020) and Dehmamy et al. (2019) argue that existing works only consider the expressivity when GNNs have infinite layers and units. Their work investigates the representation power of GNNs with finite depth and width. Oono and Suzuki (2020) discuss the asymptotic behaviors of GNNs as the model deepens and model them as dynamic systems.

#### 7.1.4. *Invariance*

As there are no node orders in graphs, the output embeddings of GNNs are supposed to be permutation-invariant or equivariant to the input features. Maron et al. (2019a) characterize permutation-invariant or equivariant linear layers to build invariant GNNs. Maron et al. (2019b) further prove the result that the universal invariant GNNs can be obtained with higher-order tensorization. Keriven and Peyré (2019) provide an alternative proof and extend this conclusion to the equivariant case. Chen et al. (2019) build connections between permutation-invariance and graph isomorphism testing. To prove their equivalence, Chen et al. (2019) leverage sigma-algebra to describe the expressivity of GNNs.

#### 7.1.5. *Transferability*

A deterministic characteristic of GNNs is that the parameterization is untied with graphs, which suggests the ability to transfer across graphs (so-called transferability) with performance guarantees. Levie et al. (2019) investigate the transferability of spectral graph filters, showing that such filters are able to transfer on graphs in the same domain. Ruiz et al. (2020) analyze GNN behaviour on graphons. Graphon refers to the limit of a sequence of graphs, which can also be seen as a generator for dense graphs. The authors conclude that GNNs are transferable across graphs obtained deterministically from the same graphon with different sizes.

#### 7.1.6. *Label efficiency*

(Semi-) Supervised learning for GNNs needs a considerable amount of labeled data to achieve a satisfying performance. Improving the label efficiency has been studied in the perspective of active learning, in which informative nodes are actively selected to be labeled by an oracle to train the GNNs. Cai et al. (2017), Gao et al. (2018b), Hu et al. (2020c) demonstrate that by selecting the informative nodes such as the high-degree nodes and uncertain nodes, the labeling efficiency can be dramatically improved.

### 7.2. *Empirical aspect*

Besides theoretical analysis, empirical studies of GNNs are also required for better comparison and evaluation. Here we include several empirical studies for GNN evaluation and benchmarks.

#### 7.2.1. *Evaluation*

Evaluating machine learning models is an essential step in research. Concerns about experimental reproducibility and replicability have been raised over the years. Whether and to what extent do GNN models work? Which parts of the models contribute to the final performance? To investigate such fundamental questions, studies about fair evaluation strategies are urgently needed.

On semi-supervised node classification task, Shchur et al. (2018a) explore how GNN models perform under same training strategies and hyperparameter tune. Their works concludes that different dataset splits lead to dramatically different rankings of models. Also, simple models could outperform complicated ones under proper settings. Errica et al. (2020) review several graph classification models and point out that they are compared inproperly. Based on rigorous evaluation, structural information turns up to not be fully exploited for graph classification. You et al. (2020) discuss the architectural designs of GNN models, such as the number of layers and the aggregation function. By a huge amount of experiments, this work provides comprehensive guidelines for GNN designation over various tasks.

#### 7.2.2. *Benchmarks*

High-quality and large-scale benchmark datasets such as ImageNet are significant in machine learning research. However in graph learning, widely-adopted benchmarks are problematic. For example, most node classification datasets contain only 3000 to 20,000 nodes, which are small compared with real-world graphs. Furthermore, the experimental protocols across studies are not unified, which is hazardous to the literature. To mitigate this issue, Dwivedi et al. (2020), Hu et al. (2020d) provide scalable and reliable benchmarks for graph learning. Dwivedi et al. (2020) build medium-scale benchmark datasets in multiple domains and tasks, while OGB (Hu et al., 2020d) offers large-scale datasets. Furthermore, both works evaluate current GNN models and provide leaderboards for further comparison.

## 8. Applications

Graph neural networks have been explored in a wide range of domains across supervised, semi-supervised, unsupervised and reinforcement learning settings. In this section, we generally group the applications in two scenarios: (1) Structural scenarios where the data has explicit relational structure. These scenarios, on the one hand, emerge from scientific researches, such as graph mining, modeling physical systems and chemical systems. On the other hand, they rise from industrial applications such as knowledge graphs, traffic networks and recommendation systems. (2) Non-structural scenarios where the relational structure is implicit or absent. These scenarios generally include image (computer vision) and text (natural language processing), which are two of the most actively developing branches of AI researches. A simple illustration of these applications is in Fig. 6. Note that we only list several representative applications instead of providing an exhaustive list. The summary of the applications could be found in Table 3.

### 8.1. *Structural scenarios*

In the following subsections, we will introduce GNNs' applications in structural scenarios, where the data are naturally performed in the graph structure.

#### 8.1.1. *Graph mining*

The first application is to solve the basic tasks in graph mining. Generally, graph mining algorithms are used to identify useful structures for downstream tasks. Traditional graph mining challenges include frequent sub-graph mining, graph matching, graph classification, graph clustering, etc. Although with deep learning, some downstream tasks can be directly solved without graph mining as an intermediate step, the basic challenges are worth being studied in the GNNs' perspective.

**Graph Matching.** The first challenge is graph matching. Traditional methods for graph matching usually suffer from high computational complexity. The emergence of GNNs allows researchers to capture the structure of graphs using neural networks, thus offering another solution to the problem. Riba et al. (2018) propose a siamese MPNN model to learn the graph editing distance. The siamese framework is two parallel MPNNs with the same structure and weight sharing, The training objective is to embed a pair of graphs with small editing distance into close latent space. Li et al. (2019b) design similar methods while experiments on more real-world scenario such as similarity search in control flow graph.

**Graph Clustering.** Graph clustering is to group the vertices of a graph into clusters based on the graph structure and/or node attributes. Various works (Zhang et al., 2019c) in node representation learning are developed and the representation of nodes can be passed to traditional clustering algorithms. Apart of learning node embeddings, graph pooling (Ying et al., 2018b) can be seen as a kind of clustering. More recently, Tsitsulin et al. (2020) directly target at the clustering task. They study the desirable property of a good graph clustering method and propose to optimize the spectral modularity, which is a remarkably useful graph clustering metric.

#### *8.1.2. Physics*

Modeling real-world physical systems is one of the most fundamental aspects of understanding human intelligence. A physical system can be modeled as the objects in the system and pair-wise interactions between objects. Simulation in the physical system requires the model to learn the law of the system and make predictions about the next state of the system. By modeling the objects as nodes and pair-wise interactions as edges, the systems can be simplified as graphs. For example, in particle systems, particles can interact with each other via multiple interactions, including collision (Hoshen, 2017), spring connection, electromagnetic force (Kipf et al., 2018), etc., where particles are seen as nodes and interactions are seen as edges. Another example is the robotic system, which is formed by multiple bodies (e.g., arms, legs) connected with joints. The bodies and joints can be seen as nodes and edges, respectively. The model needs to infer the next state of the bodies based on the current state of the system and the principles of physics.

Before the advent of graph neural networks, works process the graph representation of the systems using the available neural blocks. Interaction Networks (Battaglia et al., 2016) utilizes MLP to encode the incidence matrices of the graph. CommNet (Sukhbaatar Ferguset al., 2016) performs nodes updates using the nodes' previous representations and the average of all nodes' previous representations. VAIN (Hoshen, 2017) further introduces the attention mechanism. VIN (Watters et al., 2017) combines CNNs, RNNs and IN (Battaglia et al., 2016).

The emergence of GNNs let us perform GNN-based reasoning about objects, relations, and physics in a simplified but effective way. NRI (Kipf et al., 2018) takes the trajectory of objects as input and infers an explicit interaction graph, and learns a dynamic model simultaneously. The interaction graphs are learned from former trajectories, and trajectory predictions are generated from decoding the interaction graphs.

Sanchez et al. (2018) propose a Graph Network-based model to encode the graph formed by bodies and joints of a robotic system. They further learn the policy of stably controlling the system by combining GNs with Reinforcement learning.

#### *8.1.3. Chemistry and biology*

**Molecular Fingerprints.** Molecular fingerprints serve as a way to encode the structure of molecules. The simplest fingerprint can be a one-hot vector, where each digit represents the existence or absence of a particular substructure. These fingerprints can be used in molecule searching, which is a core step in computer-aided drug design. Conventional molecular fingerprints are hand-made and fixed (e.g., the one-hot vector). However, molecules can be naturally seen as graphs, with atoms being the nodes and chemical-bonds being the edges. Therefore, by applying GNNs to molecular graphs, we can obtain better fingerprints.

Duvenaud et al. (2015) propose neural graph fingerprints (Neural FPs), which calculate substructure feature vectors via GCNs and sum to get overall representations. Kearnes et al. (2016) explicitly model atom and atom pairs independently to emphasize atom interactions. It introduces edge representation $\mathbf{e}^t_{uv}$ instead of aggregation function, i.e. $\mathbf{h}^t_{\mathcal{N}_v} = \sum_{u \in \mathcal{N}(v)} \mathbf{e}^t_{uv}$.

**Chemical Reaction Prediction.** Chemical reaction product prediction is a fundamental issue in organic chemistry. Graph Transformation Policy Network (Do et al., 2019) encodes the input molecules and generates an intermediate graph with a node pair prediction network and a policy network.

**Protein Interface Prediction.** Proteins interact with each other using the interface, which is formed by the amino acid residues from each participating protein. The protein interface prediction task is to determine whether particular residues constitute part of a protein. Generally, the prediction for a single residue depends on other neighboring residues. By letting the residues to be nodes, the proteins can be represented as graphs, which can leverage the GNN-based machine learning algorithms. Fout et al. (2017) propose a GCN-based method to learn ligand and receptor protein residue representation and to merge them for pair-wise classification. MR-GNN (Xu et al., 2019d) introduces a multi-resolution approach to extract and summarize local and global features for better prediction.

**Biomedical Engineering.** With Protein-Protein Interaction Network, Rhee et al. (2018) leverage graph convolution and relation network for breast cancer subtype classification. Zitnik et al. (2018) also suggest a GCN-based model for polypharmacy side effects prediction. Their work models the drug and protein interaction network and separately deals with edges in different types.

#### *8.1.4. Knowledge graph*

The knowledge graph (KG) represents a collection of real-world entities and the relational facts between pairs of the entities. It has wide application, such as question answering, information retrieval and knowledge guided generation. Tasks on KGs include learning low-dimensional embeddings which contain rich semantics for the entities and relations, predicting the missing links between entities, and multi-hop reasoning over the knowledge graph. One line of research treats the graph as a collection of triples, and proposes various kinds of loss functions to distinguish the correct triples and false triples (Bordes et al., 2013). The other line leverages the graph nature of KG, and uses GNN-based methods for various tasks. When treated as a graph, KG can be seen as a heterogeneous graph. However, unlike other heterogeneous graphs such as social networks, the logical relations are of more importance than the pure graph structure.

R-GCN (Schlichtkrull et al., 2018) is the first work to incorporate GNNs for knowledge graph embedding. To deal with various relations, R-GCN proposes relation-specific transformation in the message passing steps. Structure-Aware Convolutional Network (Shang et al., 2019) combines a GCN encoder and a CNN decoder together for better knowledge representations.

A more challenging setting is knowledge base completion for out-of-knowledge-base (OOKB) entities. The OOKB entities are unseen in the training set, but directly connect to the observed entities in the training set. The embeddings of OOKB entities can be aggregated from the observed entities. Hamaguchi et al. (2017) use GNNs to solve the problem, which achieve satisfying performance both in the standard KBC setting and the OOKB setting.

Besides knowledge graph representation learning, Wang et al. (2018b) utilize GCN to solve the cross-lingual knowledge graph alignment problem. The model embeds entities from different languages into a unified embedding space and aligns them based on the embedding similarity. To align large-scale heterogeneous knowledge graphs, OAG (Zhang et al., 2019d) uses graph attention networks to model various types of entities. With representing entities as their surrounding subgraphs, Xu et al. (2019c) transfer the entity alignment problem to a graph matching problem and then solve it by graph matching networks.

#### *8.1.5. Generative models*

Generative models for real-world graphs have drawn significant attention for their important applications including modeling social interactions, discovering new chemical structures, and constructing knowledge graphs. As deep learning methods have powerful ability to learn the implicit distribution of graphs, there is a surge in neural graph generative models recently.

NetGAN (Shchur et al., 2018b) is one of the first work to build neural graph generative model, which generates graphs via random walks. It transforms the problem of graph generation to the problem of walk generation which takes the random walks from a specific graph as input and trains a walk generative model using GAN architecture. While the generated graph preserves important topological properties of the original graph, the number of nodes is unable to change in the generating process, which is as same as the original graph. GraphRNN (You et al., 2018b) manages to generate the adjacency matrix of a graph by generating the adjacency vector of each node step by step, which can output networks with different numbers of nodes. Li et al. (2018d) propose a model which generates edges and nodes sequentially and utilizes a graph neural network to extract the hidden state of the current graph which is used to decide the action in the next step during the sequential generative process. GraphAF (Shi et al., 2020) also formulates graph generation as a sequential decision process. It combines the flow-based generation with the autogressive model. Towards molecule generation, it also conducts validity check of the generated molecules using existing chemical rules after each step of generation.

Instead of generating graph sequentially, other works generate the adjacency matrix of graph at once. MolGAN (De Cao and Kipf, 2018) utilizes a permutation-invariant discriminator to solve the node variant problem in the adjacency matrix. Besides, it applies a reward network for RL-based optimization towards desired chemical properties. What's more, Ma et al. (2018) propose constrained variational auto-encoders to ensure the semantic validity of generated graphs. And, GCPN (You et al., 2018a) incorporates domain-specific rules through reinforcement learning. GNF (Liu et al., 2019) adapts normalizing flow to the graph data. Normalizing flow is a kind of generative model which uses a invertable mapping to transform observed data into latent vector space. Transforming from the latent vector back into the observed data using the inverse matrix serves as the generating process. GNF combines normalizing flow with a permutation-invariant graph auto-encoder to take graph structured data as the input and generate new graphs at the test time. Graphite (Grover et al., 2019) integrates GNN into variational auto-encoders to encode the graph structure and features into latent variables. More specifically, it uses isotropic Gaussian as the latent variables and then uses iterative refinement strategy to decode from the latent variables.

#### 8.1.6. *Combinatorial optimization*

Combinatorial optimization problems over graphs are set of NP-hard problems which attract much attention from scientists of all fields. Some specific problems like traveling salesman problem (TSP) and minimum spanning trees (MST) have got various heuristic solutions. Recently, using a deep neural network for solving such problems has been a hotspot, and some of the solutions further leverage graph neural network because of their graph structure.

Bello et al. (2017) first propose a deep-learning approach to tackle TSP. Their method consists of two parts: a Pointer Network (Vinyals et al., 2015b) for parameterizing rewards and a policy gradient (Sutton and Barto, 2018) module for training. This work has been proved to be comparable with traditional approaches. However, Pointer Networks are designed for sequential data like texts, while order-invariant encoders are more appropriate for such work.

Khalil et al. (2017), Kool et al. (2019) improve the above method by including graph neural networks. The former work first obtains the node embeddings from structure2vec (Dai et al., 2016), then feed them into a Q-learning module for making decisions. The latter one builds an attention-based encoder-decoder system. By replacing reinforcement learning module with an attention-based decoder, it is more efficient for training. These works achieve better performances than previous algorithms, which prove the representation power of graph neural networks. More generally, Gasse et al. (2019) represent the state of a combinatorial problem as a bipartite graph and utilize GCN to encode it.

For specific combinatorial optimization problems, Nowak et al. (2018) focus on Quadratic Assignment Problem i.e. measuring the similarity of two graphs. The GNN based model learns node embeddings for each graph independently and matches them using attention mechanism. This method offers intriguingly good performance even in regimes where standard relaxation-based techniques appear to suffer. Zheng et al. (2020a) use a generative graph neural network to model the DAG-structure learning problem, which is also a combinatorial optimization and NP-hard problem. NeuroSAT (Selsam et al., 2019) learns a message passing neural network to classify the satisfiability of SAT problem. It proves that the learned model can generalize to novel distributions of SAT and other problems which can be converted to SAT.

Unlike previous works which try to design specific GNNs to solve combinatorial problems, Sato et al. (2019) provide a theoretical analysis of GNN models on these problems. It establishes connections between GNNs and the distributed local algorithms which is a group of classical algorithms on graphs for solving these problems. Moreover, it demonstrates the optimal approximation ratios to the optimal solutions that the most powerful GNN can reach. It also proves that most of existing GNN models cannot exceed this upper bound. Furthermore, it adds coloring to the node feature to improve the approximation ratios.

#### 8.1.7. *Traffic networks*

Predicting traffic states is a challenging task since traffic networks are dynamic and have complex dependencies. Cui et al. (2018b) combine GNNs and LSTMs to capture both spatial and temporal dependencies. STGCN (Yu et al., 2018) constructs ST-Conv blocks with spatial and temporal convolution layers, and applies residual connection with bottleneck strategies. Zheng et al. (2020b), Guo et al. (2019) both incorporate attention mechanism to better model spatial temporal correlation.

#### 8.1.8. *Recommendation systems*

User-item interaction prediction is one of the classic problems in recommendation. By modeling the interaction as a graph, GNNs can be utilized in this area. GC-MC (van den Berg et al., 2017) firstly applies GCN on user-item rating graphs to learn user and item embeddings. To efficiently adopt GNNs in web-scale scenarios, PinSage (Ying et al., 2018a) builds computational graphs with weighted sampling strategy for the bipartite graph to reduce repeated computation.

Social recommendation tries to incorporate user social networks to enhance recommendation performance. GraphRec (Fan et al., 2019) learns user embeddings from both item side and user side. Wu et al. (2019c) go beyond static social effects. They attempt to model homophily and influence effects by dual attentions.

#### 8.1.9. *Other Applications in structural scenarios*

Because of the ubiquity of graph-structured data, GNNs have been applied to a larger variety of tasks than what we have introduced above. We list more scenarios very briefly. In financial market, GNNs are used to model the interaction between different stocks to predict the future trends of the stocks (Matsunaga et al., 2019; Yang et al., 2019; Chen et al., 2018c; Li et al., 2020). Kim et al. (2019) also predict the market index movement by formulating it as a graph classification problem. In Software-Defined Networks (SDN), GNNs are used to optimize the routing performance (Rusek et al., 2019). In Abstract Meaning Representation (AMR) graph to Text generation tasks, Song et al. (2018a), Beck et al. (2018) use GNNs to encode the graph representation of the abstract meaning.

### 8.2. *Non-structural scenarios*

In this section we will talk about applications on non-structural scenarios. Generally, there are two ways to apply GNNs on non-structural scenarios: (1) Incorporate structural information from other domains to improve the performance, for example using information from knowledge graphs to alleviate the zero-shot problems in image tasks; (2) Infer or assume the relational structure in the task and then apply the model to solve the problems defined on graphs, such as the method in (Zhang et al., 2018d) which models text into graphs. Common non-structure scenarios include image, text, and programming source code (Allamanis et al., 2018; Li et al., 2016). However, we only give detailed introduction to the first two scenarios.

### 8.2.1. Image

**Few(Zero)-shot Image Classification.** Image classification is a very basic and important task in the field of computer vision, which attracts much attention and has many famous datasets like ImageNet (Russakovsky et al., 2015). Recently, **zero-shot** and **few-shot learning** become more and more popular in the field of image classification. In $N$-shot learning, to make predictions for the test data samples in some classes, only $N$ training samples in the same classes are provided in the training set. Thereby, few-shot learning restricts $N$ to be small, and zero-shot requires $N$ to be 0. Models must learn to generalize from the limited training data to make new predictions for testing data. Graph neural networks, on the other hand, can assist the image classification system in these challenging scenarios.

First, knowledge graphs can be used as extra information to guide zero-shot recognition classification (Wang et al., 2018d; Kampffmeyer et al., 2019). Wang et al. (2018d) make the visual classifiers learn not only from the visual input but also from word embeddings of the categories' names and their relationships to other categories. A knowledge graph is developed to help connect the related categories, and they use a 6-layer GCN to encode the knowledge graph. As the over-smoothing effect happens when the graph convolution architecture becomes deep, the 6-layer GCN used in (Wang et al., 2018d) will wash out much useful information in the representation. To solve the smoothing problem, Kampffmeyer et al. (2019) use a single layer GCN with a larger neighborhood which includes both one-hop and multi-hop nodes in the graph. And it is proven effective in building a zero-shot classifier from existing ones. As most knowledge graphs are large for reasoning, Marino et al. (2017) select some related entities to build a sub-graph based on the result of object detection and apply GGNN to the extracted graph for prediction. Besides, Lee et al. (2018b) also leverage the knowledge graph between categories. It further defines three types of relations between categories: super-subordinate, positive correlation, and negative correlation and propagates the confidence of relation labels in the graph directly.

Except for the knowledge graph, the similarity between images in the dataset is also helpful for the few-shot learning (Garcia and Bruna, 2018). Garcia and Bruna (2018) build a weighted fully-connected image network based on the similarity and do message passing in the graph for few-shot recognition.

**Visual Reasoning.** Computer-vision systems usually need to perform reasoning by incorporating both spatial and semantic information. So it is natural to generate graphs for reasoning tasks.

A typical visual reasoning task is visual question answering (VQA). In this task, a model needs to answer the questions about an image given the text description of the questions. Usually, the answer lies in the spatial relations among objects in the image. Teney et al. (2017) construct an image scene graph and a question syntactic graph. Then they apply GGNN to train the embeddings for predicting the final answer. Despite spatial connections among objects, Norcliffebrown et al. (2018) build the relational graphs conditioned on the questions. With knowledge graphs, Wang et al. (2018c), Narasimhan et al. (2018) can perform finer relation exploration and more interpretable reasoning process.

Other applications of visual reasoning include object detection, interaction detection, and region classification. In object detection (Hu et al., 2018; Gu et al., 2018), GNNs are used to calculate RoI features. In interaction detection (Qi et al., 2018; Jain et al., 2016), GNNs are message-passing tools between humans and objects. In region classification (Chen et al., 2018d), GNNs perform reasoning on graphs that connects regions and classes.

**Semantic Segmentation.** Semantic segmentation is a crucial step towards image understanding. The task here is to assign a unique label (or category) to every single pixel in the image, which can be considered as a dense classification problem. However, regions in images are often not grid-like and need non-local information, which leads to the failure of traditional CNN. Several works utilize graph-structured data to handle it.

Liang et al. (2016) use Graph-LSTM to model long-term dependency together with spatial connections by building graphs in the form of distance-based superpixel map and applying LSTM to propagate neighborhood information globally. Subsequent work improves it from the perspective of encoding hierarchical information (Liang et al., 2017).

Furthermore, 3D semantic segmentation (RGBD semantic segmentation) and point clouds classification utilize more geometric information and therefore are hard to model by a 2D CNN. Qi et al. (2017b) construct a k-nearest neighbor (KNN) graph and use a 3D GNN as the propagation model. After unrolling for several steps, the prediction model takes the hidden state of each node as input and predicts its semantic label. As there are always too many points in point clouds classification task, Landrieu and Simonovsky (2018) solve large-scale 3D point clouds segmentation by building superpoint graphs and generating embeddings for them. To classify supernodes, Landrieu and Simonovsky (2018) leverage GGNN and graph convolution. Wang et al. (2018e) propose to model point interactions through edges. They calculate edge representation vectors by feeding the coordinates of its terminal nodes. Then node embeddings are updated by edge aggregation.

### 8.2.2. Text

The graph neural networks could be applied to several tasks based on texts. It could be applied to both sentence-level tasks (e.g. text classification) as well as word-level tasks (e.g. sequence labeling). We list several major applications on text in the following.

**Text Classification.** Text classification is an important and classical problem in natural language processing. Traditional text classification uses bag-of-words features. However, representing a text as a graph of words can further capture semantics between non-consecutive and long distance words (Peng et al., 2018). Peng et al. (2018) use a graph-CNN based deep learning model to first convert texts to graph-of-words, and then use graph convolution operations in (Niepert et al., 2016) to convolve the word graph. Zhang et al. (2018d) propose the Sentence LSTM to encode text. They view the whole sentence as a single state, which consists of sub-states for individual words and an overall sentence-level state. They use the global sentence-level representation for classification tasks. These methods either view a document or a sentence as a graph of word nodes. Yao et al. (2019) regard the documents and words as nodes to construct the corpus graph and use the Text GCN to learn embeddings of words and documents. Sentiment classification could also be regarded as a text classification problem and a Tree-LSTM approach is proposed by (Tai et al., 2015).

**Sequence Labeling.** Given a sequence of observed variables (such as words), sequence labeling is to assign a categorical label for each variable. Typical tasks include POS-tagging, where we label the words in a sentence by their part-of-speech, and Named Entity Recognition (NER), where we predict whether each word in a sentence belongs to a part of a Named Entity. If we consider each variable in the sequence as a node and the dependencies as edges, we can utilize the hidden state of GNNs to address the task. Zhang et al. (2018d) utilize the Sentence LSTM to label the sequence. They have conducted experiments on POS-tagging and NER tasks and achieves promising performances.

Semantic role labeling is another task of sequence labeling. Marcheggiani and Titov (2017) present a Syntactic GCN to solve the problem. The Syntactic GCN which operates on the direct graph with labeled edges is a special variant of the GCN (Kipf and Welling, 2017). It integrates edge-wise gates which let the model regulate contributions of individual dependency edges. The Syntactic GCNs over syntactic dependency trees are used as sentence encoders to learn latent feature representations of words in the sentence.

**Neural Machine Translation.** The neural machine translation (NMT) task is to translate text from source language to target language automatically using neural networks. It is usually considered as a sequence-to-sequence task. Transformer (Vaswani et al., 2017) introduces the attention mechanisms and replaces the most commonly used recurrent or convolutional layers. In fact, the Transformer assumes a fully connected graph structure between words. Other graph structure can be explored with GNNs.

One popular application of GNN is to incorporate the syntactic or semantic information into the NMT task. Bastings et al. (2017) utilize the Syntactic GCN on syntax-aware NMT tasks. Marcheggiani et al. (2018) incorporate information about the predicate-argument structure of source sentences (namely, semantic-role representations) using Syntactic GCN and compare the results between incorporating only syntactic, only semantic information and both of the information. Beck et al. (2018) utilize the GGNN in syntax-aware NMT. They convert the syntactic dependency graph into a new structure called the Levi graph (Levi, 1942) by turning the edges into additional nodes and thus edge labels can be represented as embeddings.

**Relation Extraction.** Extracting semantic relations between entities in texts helps to expand existing knowledge base. Traditional methods use CNNs or RNNs to learn entities' feature and predict the relation type for a pair of entities. A more sophisticated way is to utilize the dependency structure of the sentence. A document graph can be built where nodes represent words and edges represent various dependencies such as adjacency, syntactic dependencies and discourse relations. Zhang et al. (2018f) propose an extension of graph convolutional networks that is tailored for relation extraction and apply a pruning strategy to the input trees.

Cross-sentence N-ary relation extraction detects relations among $n$ entities across multiple sentences. Peng et al. (2017) explore a general framework for cross-sentence $n$-ary relation extraction by applying graph LSTMs on the document graphs. Song et al. (2018b) also use a graph-state LSTM model and speed up computation by allowing more parallelization.

**Event Extraction.** Event extraction is an important information extraction task to recognize instances of specified types of events in texts. This is always conducted by recognizing the event triggers and then predicting the arguments for each trigger. Nguyen and Grishman (2018) investigate a convolutional neural network (which is the Syntactic GCN exactly) based on dependency trees to perform event detection. Liu et al. (2018) propose a Jointly Multiple Events Extraction (JMEE) framework to jointly extract multiple event triggers and arguments by introducing syntactic shortcut arcs to enhance information flow to attention-based graph convolution networks to model graph information.

**Fact Verification.** Fact verification is a task requiring models to extract evidence to verify given claims. However, some claims require reasoning on multiple pieces of evidence. GNN-based methods like GEAR (Zhou et al., 2019) and KGAT (Liu et al., 2020) are proposed to conduct evidence aggregating and reasoning based on a fully connected evidence graph. Zhong et al. (2020) build an inner-sentence graph with the information from semantic role labeling and achieve promising results.

**Other Applications on Text.** GNNs can also be applied to many other tasks on text. For example, GNNs are also used in question answering and reading comprehension (Song et al., 2018c; De Cao et al., 2019; Qiu et al., 2019; Tu et al., 2019; Ding et al., 2019). Another important direction is relational reasoning, relational networks (Santoro et al., 2017), interaction networks (Battaglia et al., 2016) and recurrent relational networks (Palm et al., 2018) are proposed to solve the relational reasoning task based on text.

## 9. Open problems

Although GNNs have achieved great success in different fields, it is remarkable that GNN models are not good enough to offer satisfying solutions for any graph in any condition. In this section, we list some open problems for further researches.

**Robustness.** As a family of models based on neural networks, GNNs are also vulnerable to adversarial attacks. Compared to adversarial attacks on images or text which only focuses on features, attacks on graphs further consider the structural information. Several works have been proposed to attack existing graph models (Zügner et al., 2018; Dai et al., 2018b) and more robust models are proposed to defend (Zhu et al., 2019). We refer to (Sun et al., 2018) for a comprehensive review.

**Interpretability.** Interpretability is also an important research direction for neural models. But GNNs are also black-boxes and lack of explanations. Only a few methods (Ying et al., 2019; Baldassarre and Azizpour, 2019) are proposed to generate example-level explanations for GNN models. It is important to apply GNN models on real-world applications with trusted explanations. Similar to the fields of CV and NLP, interpretability on graphs is also an important direction to investigate.

**Graph Pretraining.** Neural network-based models require abundant labeled data and it is costly to obtain enormous human-labeled data. Self-supervised methods are proposed to guide models to learn from unlabeled data which is easy to obtain from websites or knowledge bases. These methods have achieved great success in the area of CV and NLP with the idea of pretraining (Krizhevsky et al., 2012; Devlin et al., 2019). Recently, there have been works focusing on pretraining on graphs (Qiu et al., 2020; Hu et al., 2020b, 2020e; Zhang et al., 2020), but they have different problem settings and focus on different aspects. This field still has many open problems requiring research efforts, such as the design of the pretraining tasks, the effectiveness of existing GNN models on learning structural or feature information, etc.

**Complex Graph Structures.** Graph structures are flexible and complex in real life applications. Various works are proposed to deal with complex graph structures such as dynamic graphs or heterogeneous graphs as we have discussed before. With the rapid development of social networks on the Internet, there are certainly more problems, challenges and application scenarios emerging and requiring more powerful models.

## 10. Conclusion

Over the past few years, graph neural networks have become powerful and practical tools for machine learning tasks in graph domain. This progress owes to advances in expressive power, model flexibility, and training algorithms. In this survey, we conduct a comprehensive review of graph neural networks. For GNN models, we introduce its variants categorized by computation modules, graph types, and training types. Moreover, we also summarize several general frameworks and introduce several theoretical analyses. In terms of application taxonomy, we divide the GNN applications into structural scenarios, non-structural scenarios, and other scenarios, then give a detailed review for applications in each scenario. Finally, we suggest four open problems indicating the major challenges and future research directions of graph neural networks, including robustness, interpretability, pretraining and complex structure modeling.

## Declaration of competing interest

The authors declare that they have no known competing financial interests or personal relationships that could have appeared to influence the work reported in this paper.

## Acknowledgements

This work is supported by the National Key Research and Development Program of China (No. 2018YFB1004503), the National Natural Science Foundation of China (NSFC No.61772302) and the Beijing Academy of Artificial Intelligence (BAAI). This work is also supported by 2019 Tencent Marketing Solution Rhino-Bird Focused Research Program

FR201908.

## Appendix A. Datasets

Many tasks related to graphs are released to test the performance of various graph neural networks. Such tasks are based on the following commonly used datasets. We list the datasets in Table A.4.

**Table A.4**
Datasets commonly used in tasks related to graph.

| Field | Datasets |
|---|---|
| Citation Networks | Pubmed (Yang et al., 2016) Cora (Yang et al., 2016) Citeseer (Yang et al., 2016) DBLP (Tang et al., 2008) |
| Bio-chemical Graphs | MUTAG (Debnath et al., 1991) NCI-1 (Wale et al., 2008) PPI (Zitnik and Leskovec, 2017) D&D (Dobson and Doig, 2003) PROTEIN (Borgwardt et al., 2005) PTC (Toivonen et al., 2003) |
| Social Networks | Reddit (Hamilton et al., 2017c) BlogCatalog (Zafarani and Liu, 2009) |
| Knowledge Graphs | FB13 (Socher et al., 2013) FB15K (Bordes et al., 2013) FB15K237 (Toutanova et al., 2015) WN11 (Socher et al., 2013) WN18 (Bordes et al., 2013) WN18RR (Dettmers et al., 2018) |

There are also a broader range of open source datasets repository which contains more graph datasets. We list them in Table A.5.

**Table A.5**
Popular graph learning dataset collections.

| Repository | Introduction | Link |
|---|---|---|
| Network Repository | A scientific network data repository interactive visualization and mining tools. | http://networkrepository.com |
| Graph Kernel Datasets | Benchmark datasets for graph kernels. | https://ls11-www.cs.tu-dortmund.de/staff/morris/graphkerneldatasets |
| Relational Dataset Repository | To support the growth of relational machine learning | https://relational.fit.cvut.cz/ |
| Stanford Large Network Dataset Collection | The SNAP library is developed to study large social and information networks. | https://snap.stanford.edu/data/ |
| Open Graph Benchmark | Open Graph Benchmark (OGB) is a collection of benchmark datasets, data-loaders and evaluators for graph machine learning in PyTorch. | https://ogb.stanford.edu |

## Appendix B. Implementations

We first list several platforms that provide codes for graph computing in Table B.6.

**Table B.6**
Popular platforms for graph computing.

| Platform | Link | Reference |
|---|---|---|
| PyTorch Geometric | https://github.com/rusty1s/pytorch_geometric | Fey and Lenssen (2019) |
| Deep Graph Library | https://github.com/dmlc/dgl | Wang et al. (2019b) |
| AliGraph | https://github.com/alibaba/aligraph | Zhu et al. (2019a) |
| GraphVite | https://github.com/DeepGraphLearning/graphvite | Zhu et al. (2019b) |
| Paddle Graph Learning | https://github.com/PaddlePaddle/PGL | |
| Euler | https://github.com/alibaba/euler | |
| Plato | https://github.com/tencent/plato | |
| CogDL | https://github.com/THUDM/cogdl/ | |
| OpenNE | https://github.com/thunlp/OpenNE/tree/pytorch | |

Next we list the hyperlinks of the current open source implementations of some famous GNN models in Table B.7:

**Table B.7**
Source code of the models mentioned in the survey.

| Model | Link |
|---|---|
| GGNN (2015) | https://github.com/yujiali/ggnn |
| Neurals FPs (2015) | https://github.com/HIPS/neural-fingerprint |
| ChebNet (2016) | https://github.com/mdeff/cnn_graph |
| DNGR (2016) | https://github.com/ShelsonCao/DNGR |
| SDNE (2016) | https://github.com/suanrong/SDNE |
| GAE (2016) | https://github.com/limaosen0/Variational-Graph-Auto-Encoders |
| DRNE (2016) | https://github.com/tadpole/DRNE |
| Structural RNN (2016) | https://github.com/asheshjain399/RNNexp |
| DCNN (2016) | https://github.com/jcatw/dcnn |
| GCN (2017) | https://github.com/tkipf/gcn |
| CayleyNet (2017) | https://github.com/amoliu/CayleyNet |
| GraphSage (2017) | https://github.com/williamleif/GraphSAGE |
| GAT (2017) | https://github.com/PetarV-/GAT |
| CLN (2017) | https://github.com/trangptm/Column_networks |
| ECC (2017) | https://github.com/mys007/ecc |
| MPNNs (2017) | https://github.com/brain-research/mpnn |
| MoNet (2017) | https://github.com/pierrebaque/GeometricConvolutionsBench |

(continued on next column)

**Table B.7** (*continued*)

| Model | Link |
|---|---|
| JK-Net (2018) | https://github.com/ShinKyuY/Representation_Learning_on_Graphs_with_Jumping_Knowledge_Networks |
| SSE (2018) | https://github.com/Hanjun-Dai/steady_state_embedding |
| LGCN (2018) | https://github.com/divelab/lgcn/ |
| FastGCN (2018) | https://github.com/matenure/FastGCN |
| DiffPool (2018) | https://github.com/RexYing/diffpool |
| GraphRNN (2018) | https://github.com/snap-stanford/GraphRNN |
| MolGAN (2018) | https://github.com/nicola-decao/MolGAN |
| NetGAN (2018) | https://github.com/danielzuegner/netgan |
| DCRNN (2018) | https://github.com/liyaguang/DCRNN |
| ST-GCN (2018) | https://github.com/yysijie/st-gcn |
| RGCN (2018) | https://github.com/tkipf/relational-gcn |
| AS-GCN (2018) | https://github.com/huangwb/AS-GCN |
| DGCN (2018) | https://github.com/ZhuangCY/DGCN |
| GaAN (2018) | https://github.com/jennyzhang0215/GaAN |
| DGI (2019) | https://github.com/PetarV-/DGI |
| GraphWaveNet (2019) | https://github.com/nnzhan/Graph-WaveNet |
| HAN (2019) | https://github.com/Jhy1993/HAN |

As the research filed grows rapidly, we recommend our readers the paper list published by our team, GNNPapers (https://github.com/thunlp/gnnpapers), for recent papers.

# Erratum

We are sorry for the mistakes we've made unconsciously in the paper and here are the errata. If you have any further questions, please send us an email.

- On Page 58 and 68 the citation "Huang et al. (2020)" should be corrected to "Skarding et al. (2021)". The corresponding reference should be corrected to "Foundations and modelling of dynamic networks using dynamic graph neural networks: A survey" by Joakim Skarding, Bogdan Gabrys and Katarzyna Musial.